\documentclass[11pt]{article}
	\UseRawInputEncoding
	\newcommand{\blind}{0}
	
    \makeatletter
    \renewcommand\section{\@startsection {section}{1}{\z@}%
                                       {-3.5ex \@plus -1ex \@minus -.2ex}%
                                       {2.3ex \@plus.2ex}%
                                       {\normalfont\fontfamily{phv}\fontsize{16}{19}\bfseries}}
    \renewcommand\subsection{\@startsection{subsection}{2}{\z@}%
                                         {-3.25ex\@plus -1ex \@minus -.2ex}%
                                         {1.5ex \@plus .2ex}%
                                         {\normalfont\fontfamily{phv}\fontsize{14}{17}\bfseries}}
    \renewcommand\subsubsection{\@startsection{subsubsection}{3}{\z@}%
                                        {-3.25ex\@plus -1ex \@minus -.2ex}%
                                         {1.5ex \@plus .2ex}%
                                         {\normalfont\normalsize\fontfamily{phv}\fontsize{14}{17}\selectfont}}
    \makeatother
	
	\usepackage{amsmath}
	\usepackage{graphicx}
	\usepackage{enumerate}
	\usepackage{xcolor}
	\usepackage{natbib} 
    \usepackage{longtable}
    \usepackage{algorithm}
    \usepackage{booktabs}
    \usepackage{algpseudocode}
    \usepackage{hyperref}
    \usepackage{dsfont}
    \usepackage{multirow, amssymb, amsmath, color,float}
	
\begin{document}
		
		\def\spacingset#1{\renewcommand{\baselinestretch}%
			{#1}\small\normalsize} \spacingset{1}
		
		\if0\blind
		{
			\title{\bf Rolling Lookahead Learning for Optimal Classification Trees}
			\author{Zeynel Batuhan Organ $^a$, Enis Kay{\i}\c{s} $^a$ and Taghi Khaniyev $^b$ \\
			$^a$ Deparment of Industrial Engineering, Ozyegin University, Istanbul, Turkey \\
             $^b$ Department of Industrial Engineering, Bilkent University, Ankara, Turkey}
			\date{}
			\maketitle
		} \fi
		
		\if1\blind
		{

             \title{\bf Rolling Lookahead Learning for Optimal Classification Trees}
			\author{Author information is purposely removed for double-blind review}
			\maketitle
			
			\medskip
		} \fi
		
	\begin{abstract}

Classification trees continue to be widely adopted in machine learning applications due to their inherently interpretable nature and scalability. We propose a rolling subtree lookahead algorithm that combines the relative scalability of the myopic approaches with the foresight of the optimal approaches in constructing trees. The limited foresight embedded in our algorithm mitigates the learning pathology observed in optimal approaches. At the heart of our algorithm lies a novel two-depth optimal binary classification tree formulation flexible to handle any loss function. We show that the feasible region of this formulation is an integral polyhedron, yielding the LP relaxation solution optimal. Through extensive computational analyses, we demonstrate that our approach outperforms optimal and myopic approaches in 808 out of 1330 problem instances, improving the out-of-sample accuracy by up to 23.6\% and 14.4\%, respectively.

	\end{abstract}
			
	\noindent%
	{\it Keywords:} Optimal classification tree, learning pathology, rolling lookahead.

	\spacingset{1.5} 

\section{Introduction} \label{s:intro}

Since their introduction by \cite{gordon_classification_1984}, decision trees (DTs) have been among the most widely used methods for predictive modeling, including classification and regression problems. Their widespread use across many application domains ranging from healthcare to finance has been attributed to the intuitive nature, computational efficiency, and interpretability of DTs \citep{bertsimas_optimal_2017}. However, with the rise of machine learning, more complex tree-based techniques such as Random Forests  \citep{breiman_random_2001} and Gradient Boosting Trees \citep{friedman2001greedy} have gained popularity due to their superior predictive performance and computational efficiency. Despite their advantages, these advanced methods are often viewed as black-box models, limiting their adoption in practical applications. In scenarios where decision-makers are not only interested in accurately predicting a certain outcome, but also in understanding why a certain prediction is made by the model, DTs have retained their appeal over black-box models \citep{rudin2019stop}.

The main idea of DTs is to cluster datapoints based on a set of split conditions orchestrated in a special (i.e., tree) structure. The accuracy of a DT hinges on the selection of these conditions, which collectively form a tree. Two main approaches exist for how the conditions are selected to construct a DT: \textit{myopic} and \textit{optimal}. 
In the myopic approach, a tree is created sequentially, with each node being split into child nodes using a condition on the value of a feature that provides the best improvement with respect to a predefined objective or loss function (such as Gini impurity or misclassification error). These child nodes are, in turn, each split into their own child nodes in the same manner, using (possibly) a different feature, while locally optimizing the objective function. This process continues iteratively until no further splits yield improvement in the objective function or a termination condition is met. Although myopic approaches are generally fast, their focus on looking only one step ahead can result in sub-optimal splits that undermine the overall accuracy of the tree.

Optimal trees, in contrast, select split conditions for all internal nodes simultaneously to optimize the objective function for the overall tree. The progress in mixed integer optimization (MIO) techniques over the recent decades has encouraged researchers to formulate this problem as an MIO model that seeks the best possible tree of a predefined depth. This approach is guaranteed to yield the most accurate tree in terms of in-sample accuracy if allowed to run to optimality. In practice, however, even for moderately sized problem instances it rarely reaches optimality within hours, due to the complexity of the underlying optimization problem. Once the optimality guarantee is removed, the MIO formulation is practically reduced to yet another computationally expensive heuristic.

Balancing accuracy and scalability is a critical consideration for constructing decision trees in practical applications with large datasets and numerous features. 
Typically, when the size of an MIO problem grows, researchers often resort to heuristic methods that offer demonstrably good results. The challenge of constructing a DT is that some of the well-known myopic heuristics have proven to be quite effective. Thus, for a proposed heuristic to have any practical merit, it must deliver consistently higher accuracy than these benchmark heuristics within a reasonable time, which we argue is attainable by decomposing the overall DT construction problem into smaller subtree constructions where the resulting MIO formulation is much easier to solve.

\subsection{Our Approach}

We propose an approach that lies between the two existing extremes of the spectrum of constructing a DT. The myopic approach is basically a one-step lookahead algorithm that optimizes locally at each node while the optimal approach can be interpreted as a $D$-step lookahead algorithm that optimizes globally for the overall tree. As an alternative to these approaches, we propose a $k$-step lookahead ($1<k\leq D$) algorithm for choosing split conditions at each node of the tree.

\begin{figure}[htb]
    \centering
    \includegraphics[width=17cm]{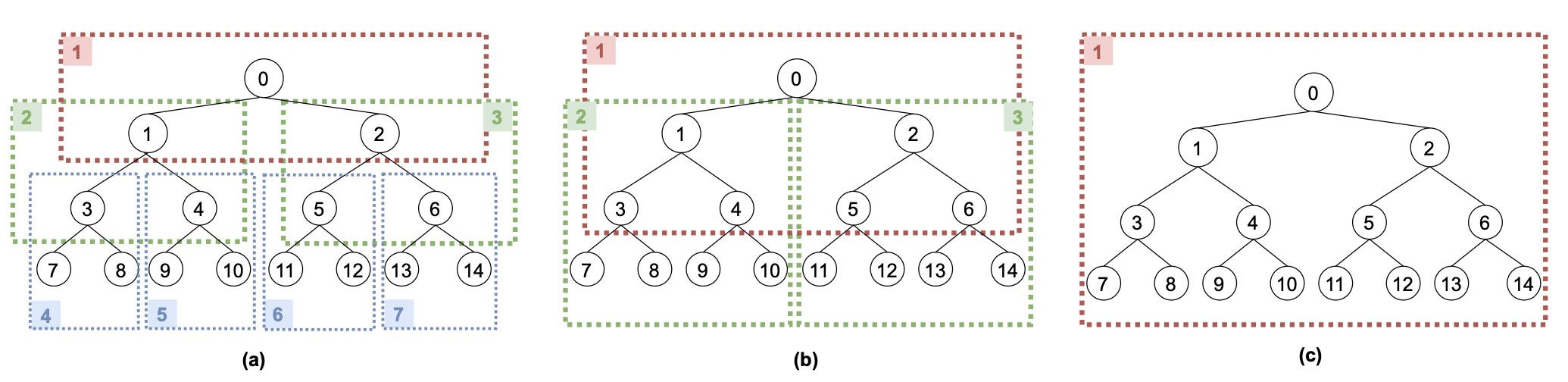}
    \caption{Visual illustration of decision tree construction methods: (a) myopic, (b) two-step lookahead, (c) optimal}
    \label{fig:approach}
\end{figure}

Figure \ref{fig:approach} illustrates the construction of a 3-depth tree using (a) myopic, (b) two-step lookahead, and (c) optimal approaches. In our proposed approach (Figure \ref{fig:approach}(b)), we first construct a 2-depth tree at the root node (Node 0, highlighted with a red dashed square). Then, the split condition on the first level is fixed and the node is split into two child nodes (Nodes 1 and 2) based on this condition. Then, we continue with the child nodes (Nodes 1 and 2). For each child node, we again construct another 2-depth tree starting from these nodes (highlighted with green dashed squares). This process continues until we reach the predefined depth $D$ ($D=3$, in this example) for the overall tree. To compare, the myopic approach solves 7 small one-step lookahead optimization problems, our approach solves 3 moderate size two-step lookahead problems, and the optimal approach solves a single large three-step lookahead problem.


Looking $k$-steps ahead requires solving an optimal tree problem with the predefined depth of $k$, which, as discussed above, constitutes a computational bottleneck, especially for large depths. To alleviate the computational burden of the lookahead, we propose a new MIO formulation that leverages the fact that $k$ is fixed and small (e.g., $k=2$). Specifically, we first enumerate all possible $k$-tuple combinations of split conditions (referred to as \textit{leaf decision rules}) and pre-compute the objective value corresponding to each candidate leaf decision rule. We, then, devise a new MIO formulation to choose the $2^k$ leaf decision rules (one for each leaf node) that collectively form a tree and optimize the overall objective function. When all features are binary, this requires $2^k {p \choose k}$ pre-computations, where $p$ is the number of features. This enumeration is quite scalable for $k=2$ (but not so much for $k\ge3$), for most realistic instances. The proposed MIO formulation is faster than existing MIO formulations in the literature because it is formulated for a specific depth of $k=2$. Furthermore, it scales well with respect to the number of datapoints which is not true for the existing MIO formulations. Finally, the new formulation allows for the use of complex loss functions, including non-linear, non-convex ones.

\subsection{\emph{Literature}} \label{s:literature}
Decision trees are easy to interpret, yet their construction is not trivial. The literature on this subject could be categorized into two broad groups based on the solution approach employed. 

Traditionally, DTs are constructed using myopic (greedy) approaches. Some of the most acknowledged ones are classification and regression trees (CART) \citep{gordon_classification_1984}, C4.5 \citep{quinlan_c4.5:_1993}, and ID3 \citep{quinlan_induction_1986}. These methods commonly aim to find the best split condition for one node at a time, ignoring the effects of such a split on the subsequent splits. They employ a top-down algorithm that starts splitting from the root node and grows up the tree by subsequent splits at each node until a predefined depth is attained. On the other hand, the objective function that is to be optimized differs: CART typically uses Gini impurity, ID3 mainly uses entropy, and C4.5, which is an extension of ID3, uses normalized information entropy as the objective function.

Papers in the second group aim to improve the accuracy of the constructed DTs using mathematical optimization. Indeed, optimization has been used for various sub-tasks involved in creating DTs. For instance, \cite{bennett_decision_1992} proposes a linear programming formulation for determining an optimal linear split of a given node. Given a tree topology of a prespecified size,  \cite{gunluk2021optimal} proposes an MIO formulation for finding optimal trees up to depth three and effective for moderate-sized datasets. Similarly, \cite{Verwer_learning_2017} introduces another MIO formulation, which was tested for trees up to depth five, but without reporting any out-of-sample accuracy results. Both of these studies use misclassification as the objective function, employed commercial optimization software to solve the resulting formulation and demonstrate that warm starts could improve solution quality and time, particularly for medium-sized datasets. In contrast, \cite{Bennett96optimaldecision} presents a tabu search algorithm for solving the complex optimization problem. Finally, \cite{blanquero_sparsity_2020} and  \cite{blanquero_optimal_2021} introduce optimal randomized decision trees, and regularization in such trees, which formulates a continuous optimization problem to determine the best randomized splits at each node of the tree.    

Papers that use optimization to construct decision trees mainly leverage two alternative techniques: dynamic programming and MIO. Papers in the former group utilize a bottom-up approach. For example, \cite{garofalakis2003building} proposes a time and memory efficient dynamic programming-based algorithm under size or accuracy constraints. \cite{nijssen2010optimal} introduces an algorithm exploiting ideas from pattern mining to generate constraints for optimal decision tree construction. The branch-and-bound approach is employed in \cite{aglin2020learning} and subtrees of earlier iterations are stored in the cache for later reuse to speed up the algorithm. \cite{lin_generalized_2020} presents the GOSDT algorithm to address problems faced with imbalanced datasets and scalability issues when dealing with continuous features. In these papers, various linear objective functions are studied (i.e., Gini impurity is excluded). Finally, \cite{demirovic_murtree:_2020} limits the number of total nodes in the tree, but is limited to small depths (i.e., depth four) to be scalable. Beyond decision trees, dynamic programming is used in \cite{angelino2018learning} to construct the best rule lists, which are simply one-sided decision trees, while utilizing efficient data structures and branch and bound algorithms. An extension of this idea to decision tree generation is studied in \cite{hu_optimal_2019}. 

MIO techniques are increasingly being used to train globally optimal trees. \cite{bertsimas_optimal_2017} proposes an algorithm that utilizes MIO to construct optimal classification trees (OCTs). However, the proposed model does not scale well when the number of datapoints is large as it defines variables and constraints for each datapoint. To have a formulation that is largely independent of the number of datapoints, \cite{verwer_learning_2019} proposes $\it{BinOCT}$ which is reported to outperform OCT. On the other hand, $\it{BinOCT}$ sacrifices sparsity and is designed to construct complete and balanced binary decision trees that may include redundant leaves. A strong max-flow based model ($\it{FlowOCT}$) with good branch-and-bound performance, owing partly to its $\it{big-M}$ free formulation, is introduced in \cite{aghaei_learning_2020}. In the same paper, an accelerated alternative solution that utilizes Bender's decomposition, $\it{BendersOCT}$, is also presented.   

Overfitting frequently arises while training the model due to using many features and parameters \citep{wright2022optimization}. To address this challenge, formulations include a regularization parameter to control the complexity of the trained model. \cite{bertsimas_optimal_2017} and \cite{aghaei_learning_2020} minimizes the number of splits along with misclassification error, whereas \cite{hu_optimal_2019} penalizes the number of leaves. Unfortunately, these remedies may not eliminate the problem. \cite{elomaa2005look} argues that looking many steps ahead may harm out-of-sample accuracy (also known as learning pathology). Myopic approaches and global optimization approaches are at the two ends of this spectrum. Looking two steps ahead provides a good compromise between accuracy and complexity \citep{rokach2016decision,murthy_lookahead_1995}. \cite{last_avoiding_2013} and \cite{esmeir2004lookahead} propose non-optimization based procedures that utilize this limited lookahead idea.  

We propose a decision tree construction algorithm that implements a limited lookahead approach using an MIO model. Borrowing the rolling horizon idea from predictive analytics literature (see, for example, \cite{sethi1991theory, wang2015rolling}), our algorithm re-optimizes split conditions at every level by looking two steps ahead. Our model is flexible enough to handle any objective function, such as misclassification error and Gini impurity. Existing MIO formulations do not consider nonlinear and/or continuous objective functions.   

\subsection{\emph{Contributions}} \label{s:contributions}

Our contributions to the literature can be summarized as follows:

\begin{enumerate}
    \item We propose an original MIO formulation to find the optimal 2-depth tree that scales well with the number of datapoints. For example, on a large dataset with approximately 50,000 datapoints and 135 features, our formulation finds a DT with zero optimality gap under 2 minutes, whereas BendersOCT, one of the fastest among optimal approaches, reports 100\% optimality gap after 30 minutes. 
    \item Our MIO formulation can handle any objective function, including non-linear ones such as Gini impurity, by moving the computational/optimization burden of non-linearity to a pre-calculation stage. 
    \item We devise an algorithm that constructs a tree of any given depth utilizing the proposed MIO formulation iteratively via a rolling subtree technique. 
    \item Via comprehensive computational experiments, we compare the out-of-sample accuracy of our algorithm against the state-of-the-art algorithms. In the majority of the tested instances, the proposed approach outperforms optimal and myopic benchmarks and improves the out-of-sample accuracy by up to 23.6\% and 14.4\% with respect to optimal and myopic approaches, respectively. Our method mitigates the learning pathology problem associated with optimal trees but enjoys the benefits of mathematical optimization, unlike myopic approaches. 
\end{enumerate}

\subsection{\emph{Paper Structure}} \label{s:PaperStructure}

The rest of the paper is organized as follows: We begin Section \ref{s:methods} by providing details of the problem and presenting our MIO formulation for a 2-depth tree accompanied by related analytical results for the model. Next, in Section \ref{s:algo} we outline the rolling subtree algorithm to generate a DT of any given depth. We discuss the implementation of the algorithm with two alternative objective function formulations in Section \ref{s:OptCrit}. The setup and the results of the computational experiments are discussed in Section \ref{s:CompExps}. Finally, conclusions and future research directions are provided in Section \ref{s:conclusion}.

\section{Modelling Approach} \label{s:methods}
We consider the problem of learning a decision tree classifier (i.e, \emph{classification tree}) which assigns datapoints to one of the pre-defined classes $\mathcal{C}$, given a training dataset. The dataset consists of $n$ datapoints $\{(\textbf{x}_i; y_i)\}_{i=1}^{n}$, where $\textbf{x}_i$ and $y_i$ denote the $p$-dimensional feature vector and the class label of the $i^{th}$ datapoint, respectively. Hereinafter, for simplicity of exposition, we assume all features are binary-valued, similar to \cite{verwer_learning_2019}. An all-binary-features version of a dataset can be obtained via standard conversion methods. Specifically, categorical features are converted into multiple binary features via one-hot coding (one feature for each unique category). Numerical features are first divided into intervals (e.g., based on quartiles) and encoded as categorical features, which are, then, binarized.

\begin{figure}
    \centering
    \includegraphics[width=12cm]{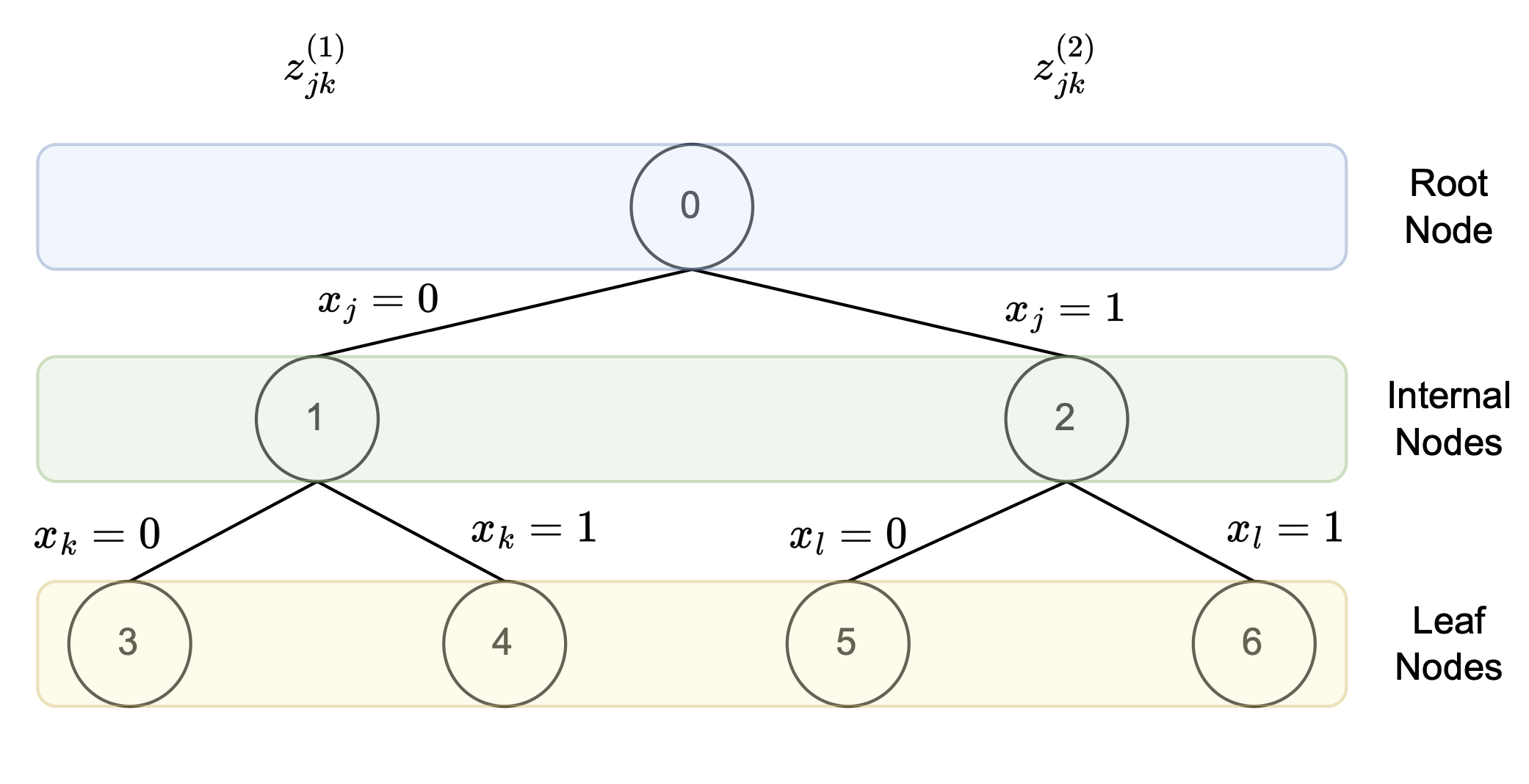}
    \caption{Illustration of notations on an example complete 2-depth tree}
    \label{fig:2_depth}
\end{figure}

The building block of our proposed approach is learning an optimal 2-depth tree for a given subset of datapoints. Figure \ref{fig:2_depth} illustrates an example of a complete 2-depth (binary) tree. Using this simple decision tree, we introduce the notation used throughout the paper. Node 0 is called the root node; Nodes 1 and 2 are called internal nodes; and Nodes 3-6 are called leaf nodes. The root node as well as each of the internal nodes contain two \textit{split conditions} (one for each child node) based on the value of a specific feature. For example, on the tree illustrated in Figure \ref{fig:2_depth}, split conditions of the root node are based on feature $j$, whereas the split conditions of Node 1 and Node 2 are based on feature $k$ and feature $l$, respectively. Unlike the root node and the internal nodes, leaf nodes do not contain any split conditions. They, instead, hold a unique class label. In a classification tree, each datapoint starts at the root node, passes through exactly one of the internal nodes at each level, and ends up at exactly one of the leaf nodes and assumes the class label of that node. Specifically, at the root node, datapoints satisfying the split condition $x_j=0$ flow left to Node 1 and those satisfying $x_j=1$ flow right to Node 2. Similarly, at Node 1 (Node 2), datapoints with $x_k=0$ ($x_l=0$) flow left to Node 3 (Node 5), and those with $x_k=1$ ($x_l=1$) flow right to Node 4 (Node 6). It is worth noting that although Figure \ref{fig:2_depth} displays a complete binary tree, the MIO model we subsequently present is applicable to incomplete/imbalanced trees as well, since any incomplete tree can be trivially represented by a complete tree. For example, if the split feature on Node 1 was also feature $j$ (instead of feature $k$), then this would imply that Node 1 is not split further, i.e., in its complete tree representation, all the datapoints in Node 1 flow into one of its children and none to the other. 

To train a 2-depth classification tree for a dataset with all-binary features essentially translates into identifying the three features ($j,k$, and $l$) to be used in the split conditions of the Nodes 0, 1, and 2, respectively, such that a predetermined loss function (e.g., misclassification error, Gini impurity, entropy, etc.) is minimized. In the absence of any other constraint/insight, this can be achieved by a brute-force enumeration of $O(p^3)$ possible combinations of features used for split conditions. That is, calculating the corresponding loss for each possible combination \emph{a priori}, and picking the one with the minimum value. The key insight that helps us reduce the number of enumerations comes from the decomposability of the tree structure. Specifically, to compute the overall loss of a decision tree, one can compute the components of the loss function contributed by each leaf node, separately, and then sum them up. Our modeling approach exploits this insight in two steps: First, all $O(p^2)$ possible split condition pairs (i.e., leaf decision rules) are enumerated for each of the leaf nodes 3-6. Then, a mathematical optimization model is devised to select the best \emph{decision rule set} (which contains exactly one decision rule for each leaf) while making sure the decision rules selected for the leaves are compatible with each other, i.e., together they form a 2-depth tree. 

As an alternative to complete enumeration, our approach breaks up the complexity into two stages: partial enumeration followed by optimization. To illustrate the reduction in the number of enumerations, consider a dataset with $p=100$ features. A complete enumeration would require $100^3 = 1,000,000$ loss function calls, whereas the partial enumeration in our approach requires $4\times100^2 = 40,000$ loss function component calls. The hope is that the reduction in the enumeration time far exceeds the time it takes to run the subsequent optimization model. 

In the partial enumeration stage, the loss function component for each possible leaf decision rule is pre-computed leading to the following parameters used in the subsequent optimization stage: $c_{jk}^{(r,s)}, \ \forall (r,s)\in\{0,1\}^{2} \text{ and } \forall j,k\in\mathcal{F}$ where $\mathcal{F}$ is the set of features. $c_{jk}^{(r,s)}$ is defined as the loss function component evaluated over the subset of datapoints for which the leaf decision rule  \{$x_j = r \text{ AND } x_k = s$\} holds. Once the loss function component for each possible leaf decision rule ($c_{jk}^{(0,0)}, c_{jk}^{(0,1)}, c_{jl}^{(1,0)}, c_{jl}^{(1,1)}$) is pre-computed, we next select the best decision rule set via an optimization model. 

We start by defining the following set of binary decision variables: 
$$ z_{jk}^{(1)} = \left\{
\begin{array}{ll}
    1, & \text{if $j^{th}$ feature is used in the split conditions of the root node} \\
        & \text{\& $k^{th}$ feature is used in the split conditions of   \underline{Node 1}}\\
    0, & o.w.
\end{array}\right.$$

$$ z_{jl}^{(2)} = \left\{
\begin{array}{ll}
    1, & \text{if $j^{th}$ feature is used in the split conditions of the root node} \\
        & \text{\& $l^{th}$ feature is used in the split conditions of   \underline{Node 2}}\\
    0, & o.w.
\end{array}\right.$$

We can now formulate the optimal 2-depth classification tree model (hereinafter, referred to as [OCT-2]) as follows:

\begin{subequations}
\begin{align}
&&\text{min} &\sum_{j \in  \mathcal{F}}\sum_{k \in  \mathcal{F}} c_{jk}^{(0,0)}z_{jk}^{(1)}+\sum_{j \in  \mathcal{F}}\sum_{k \in  \mathcal{F}} c_{jk}^{(0,1)}z_{jk}^{(1)} &+ \sum_{j \in  \mathcal{F}}\sum_{l \in  \mathcal{F}} c_{jl}^{(1,0)}z_{jl}^{(2)} &+ \sum_{j \in  \mathcal{F}}\sum_{l \in  \mathcal{F}} c_{jl}^{(1,1)}z_{jl}^{(2)} &  \\
&&\text{s.t. }&\sum_{j \in  \mathcal{F}}\sum_{k \in  \mathcal{F}}z_{jk}^{(1)} = 1&    \\
&&			&\sum_{j \in  \mathcal{F}}\sum_{l \in  \mathcal{F}}z_{jl}^{(2)} = 1&    \\
&&			&\sum_{k \in  \mathcal{F}}z_{jk}^{(1)} - \sum_{l \in  \mathcal{F}}z_{jl}^{(2)} = 0&   \ \forall j \in \mathcal{F}  \\
&&            &z_{jk}^{(1)} \in \{0,1\}&  \forall j,k \in \mathcal{F}  \\
&&            &z_{jl}^{(2)}  \in \{0,1\}&  \forall j,l \in \mathcal{F} 
\end{align}
\end{subequations}

Four components of the objective function (1a) each represent the loss value contributed by the leaf nodes 3-6. Constraint (1b) ensures that exactly one pair of features $(j,k)$ are selected for the leaf nodes 3 and 4. Similarly, constraint (1c) ensures that exactly one pair of features $(j,l)$ are selected for the leaf nodes 5 and 6. Constraints (1d) ensure that the split feature selected for the first level is the same for both the left leaves \{3, 4\} and the right leaves \{5, 6\}. 

The next two propositions show that one can relax the integrality constraints in [OCT-2] in finding the optimal solution.

\textbf{Proposition 1:} Constraint matrix of [OCT-2] is totally unimodular (TU). 

\textbf{Proof:} Let $A$ be an $m \times n$ constraint matrix of [OCT-2]. \cite{dantzig1956linear} shows that the following statements are sufficient for $A$ to be TU:
\begin{enumerate}[(i)]
    \item Every entry in $A$ is 0, +1, or −1.
    \item Every column of $A$ contains at most two non-zero (i.e., +1 or −1) entries.
    \item There exists a partition $M_1 \cup M_2 = [m]$ of the rows such that every column j with two non-zero entries satisfies: $\sum_{i \in M_1} A_{ij} = \sum_{i \in M_2} A_{ij}$. 
\end{enumerate}

Trivially (i) holds. To show (ii), observe that columns corresponding to the decision variables $z_{jk}^{(1)}$ each have a coefficient of $+1$ in the constraint (1b), a coefficient of $+1$ in the $j^{th}$ constraint of (1d), and 0 elsewhere. Similarly, columns corresponding to the decision variables $z_{jl}^{(2)}$ each have a coefficient of $+1$ in constraint (1c), a coefficient of $-1$ in the $j^{th}$ constraint of (1d), and 0 elsewhere. Finally, letting ${M_1} = \{(1b)\}$ and ${M_2} = \{(1c), (1d)\}$ proves that (iii) is also satisfied. $\blacksquare$ \\

\textbf{Proposition 2:} The feasible region of [OCT-2] is an integral polyhedron. That is, the linear programming (LP) relaxation of the [OCT-2] formulation always provides an integer optimum solution. 

\textbf{Proof:} As proven in \cite{hoffman1956integral}, if the constraint matrix ($A$) is TU and the right hand side $b$ is integral, then the polyhedron $\{x: Ax \le b\}$ is integral. Per Proposition 1, the constraint matrix $A$ of [OCT-2] is TU. Moreover, the right-hand side vector $b = (1,1,0,0,...,0)$ consists only of integers. Therefore, the feasible region of [OCT-2] is an integral polyhedron. $\blacksquare$ \\

[OCT-2] formulates a 2-depth DT construction problem by looking two steps ahead, which is known to mitigate the learning pathology that is observed for deeper lookahead approaches. Our modeling approach can be extended to 
construct an optimal 3-depth DT which is provided in the Online Appendix A.   

To mitigate the risk of overfitting, benchmark approaches (such as CART and OCT) have options to impose a minimum number of datapoints in any node. In the same spirit, our base model can also be augmented by incorporating additional constraints that impose a minimum number of datapoints to flow through each node. For example, during the partial enumeration stage, another set of parameters $(n_j, \ \forall j \in \mathcal{F})$ can be pre-computed indicating the number of datapoints that would flow into Node 1 if  feature $j$ is used as the split feature on the root node. In that case, the number of datapoints flowing into Node 2 would be $n-n_j$. Then, the following set of constraints can be added to the model to impose a minimum number of datapoints $(n_{int})$ for the internal nodes:

\begin{equation}
n_{int}\sum_{k \in  \mathcal{F}}z_{jk}^{(1)} \le \min\{n_j, n-n_j\}, \ \ \ \forall j \in \mathcal{F}. \\
\end{equation}

Similarly, if a minimum number of datapoints for the leaf nodes is to be imposed, another set of parameters $n_{jk}^{(0)} \left(n_{jl}^{(1)}\right)$, can be pre-computed indicating the number of datapoints for which the decision rule $\{x_j = 0 \text{ AND } x_k = 0\} \left(\{x_j = 1 \text{ AND } x_l = 1 \}\right)$ holds. Then, the following set of constraints can be added to the model to impose a minimum number of datapoints $(n_{leaf})$ for the leaf nodes. 
\begin{subequations}
\begin{align}
& n_{leaf}z_{jk}^{(1)} \le \min\{n_{jk}^{(0)}, n_j-n_{jk}^{(0)}\}, \ \ \ \forall j,k \in \mathcal{F}
\\
& n_{leaf}z_{jl}^{(2)} \le \min\{n_{jl}^{(1)}, n-n_j-n_{jl}^{(1)}\}, \ \ \ \forall j,l \in \mathcal{F}
\end{align}
\end{subequations}

To be able to solve the proposed MIO formulation, we first need to compute the $c_{jk}^{(r,s)}$ parameters. Although the general framework presented here is applicable to any loss function, pre-computation of the loss function components depends on the type/definition of the function used. We discuss two alternative loss functions and their implementation details in Section \ref{s:OptCrit}.

\newpage
\section{Algorithmic Approach} \label{s:algo}
The optimization model [OCT-2] presented in the previous section finds the optimal 2-depth tree, yet one may need to generate deeper trees, especially for datasets with a large number of features and datapoints. To address that challenge, we design a rolling subtree (RST) algorithm, motivated by the rolling horizon approaches employed in forecasting. 

\begin{figure}[h!]
    \centering
    \includegraphics[width=14cm]{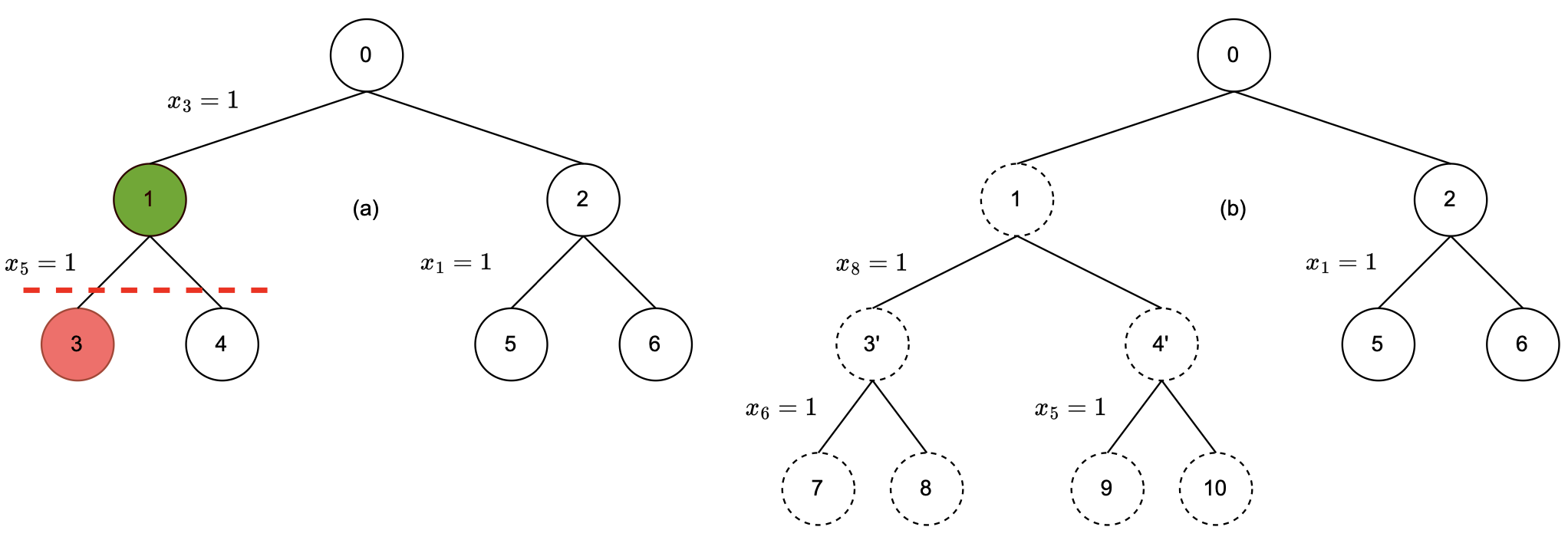}
    \caption{A simple illustration of RST algorithm}
    \label{fig:rolling}
\end{figure}

Our algorithm starts at the root node by generating an optimal 2-depth tree using all the available datapoints. Misclassified datapoints at the leaf nodes are identified, if any. Next, we identify the internal nodes that are the parents of the current leaf nodes with misclassified datapoints. Then, we reuse [OCT-2] model to generate a new subtree starting from these internal nodes, discarding all the leaf nodes that emanated in the previous step from them. The algorithm stops when there are no leaf nodes with misclassified datapoints or the maximum depth size ($D_{max}$) is achieved. The pseudo-code of our RST algorithm is presented in Algorithm \ref{algo:rst}.

An example implementation of our rolling-subtree algorithm is illustrated in Figure \ref{fig:rolling}. In this example, an optimal 2-depth tree is constructed starting from the root node (Node 0) as shown in subplot (a). Assume that Node 3 has some misclassified datapoints. Then, we select its internal node (Node 1) and find an optimal 2-depth tree starting from this node as shown in subplot (b). In this step, we discard any leaf nodes, and any split conditions thereof, emanating from Node 1. Note that the split condition has changed while generating the new 2-depth tree: The algorithm used feature 5 to separate Nodes 3 and 4, whereas feature 8 is used to separate updated nodes 3' and 4'. Since there is no misclassification in all the leaf nodes (5-10), the algorithm stops at this stage for this illustrative example. 
 
\begin{algorithm}
	\caption{RST Algorithm} 
	\begin{algorithmic}[1]
	    \State \textbf{Step 0:} Initialize the set of nodes with misclassification $\mathcal{S}=\{0\}$ and tree depth $D=0$
     	\While {$D < D_{max}$ OR $\mathcal{S}\neq \emptyset$}
                \State Select a node from the set $\mathcal{S}$ with the lowest depth (in case of ties, select the node with the smallest index)
                \State Solve the resulting [OCT-2] problem growing from the selected node 
                \State Remove the current node and add internal nodes of all resulting leaf nodes with misclassification to the set $\mathcal{S}$ 
                \State Set $D$ equal to 1 plus the smallest depth of the nodes in the set $\mathcal{S}$ 

		\EndWhile
	\end{algorithmic} 
 \label{algo:rst}
\end{algorithm}

Our algorithm inherits good characteristics of the myopic and optimal approaches while alleviating some of the concerns associated with them. Myopic approaches generally consider only one step ahead and do not have the flexibility to modify the split conditions set at earlier stages of the algorithm. On the other hand, the RST algorithm could modify split conditions at later stages during re-optimization, if necessary. Similar to the optimal approaches, our algorithm considers more than just the immediate child nodes in determining the best split condition. Constructing a deep decision tree with the optimal approach (i.e., looking $D$ steps ahead) could lead to learning pathology. To mitigate this problem, our RST algorithm constructs many intermittent 2-depth subtrees sequentially resulting in the overall tree when combined. Finally, solving 2-depth trees saves computation time, which could be extremely long to construct optimal deep trees for big datasets. 


RST algorithm requires successive calls to the [OCT-2] formulation, which may pose a potential problem. While this is true in theory, notice that as the tree grows the number of datapoints in each node decreases leading to much lower pre-computation and optimization times in the subsequent calls. Moreover, our algorithm stops splitting nodes with perfect classification. These, in turn, allow growing deeper trees within a reasonable time. Finally, although not implemented in this paper, tree growing across different branches could be run independently in parallel to further decrease computational time. As discussed in Section \ref{s:CompExps}, our algorithm requires much shorter computational times as compared to optimal trees.

\section{Optimization Criteria} \label{s:OptCrit}

Our RST algorithm is flexible to handle any loss function. In this section, we specifically discuss the implementation using two alternative loss functions: (i) \emph{misclassification error}, which is the most common function used in  optimal tree formulations due to its linear nature, and (ii) \emph{Gini impurity}, which is used in the myopic heuristic CART. The first step of the pre-computation is the same for both functions: to filter the datapoints based on the leaf decision rule they satisfy. Beyond this, the two functions are fundamentally different from each other in the way they work. Specifically, misclassification error counts the number of misclassified datapoints assuming a single class label is assigned to each leaf. Gini impurity, on the other hand, measures the likelihood that a randomly chosen datapoint is misclassified if a leaf is assigned a random class label based on the class distribution of the datapoints within that leaf  \citep{gordon_classification_1984}. 

Let $\mathcal{D}_{jk}^{(r,s)}$ and $n_{jk}^{(r,s)}$ be the subset and the number of datapoints, respectively, for which the leaf decision rule \{$x_j = r \text{ AND } x_k = s$\} holds. The misclassification error for this subset can be computed as follows:

\begin{equation}
c_{jk}^{(r,s)} = \frac{1}{n}\left(n_{jk}^{(r,s)} -  \max_{c \in \mathcal{C}} \{\sum_{i \in \mathcal{D}_{jk}^{(r,s)}} \mathds{1}_{\{y_i= c\}} \} \right) \\
\end{equation}
In equation (4), we first count the number of misclassified datapoints within a given leaf, where the class label of the leaf is determined by the simple majority rule. Next, we normalize the number of misclassified datapoints with the total number of datapoints $n$. Thus, when the individual loss function components from each leaf are summed as in the objection function (1a) of [OCT-2], we get the \emph{normalized} misclassification error of the overall tree. 

For the same subset $\mathcal{D}_{jk}^{(r,s)}$, the Gini impurity can be computed as follows:

\begin{equation}
c_{jk}^{(r,s)} =  \frac{n_{jk}^{(r,s)}}{n}\left( 1 -  \sum_{c \in \mathcal{C}}p_{c}^{2} \right) \\
\end{equation}
where 
$p_{c} = \frac{\sum_{i \in \mathcal{D}_{jk}^{(r,s)}} \mathds{1}_{\{y_i= c\}}}{n_{jk}^{(r,s)}}$ is the fraction of datapoints with class label $c$ in the subset. In equation (5), we first compute the non-normalized impurity of a leaf (the term in parenthesis), then normalize it with the term $\frac{n_{jk}^{(r,s)}}{n}$. This normalization term is needed to compute the Gini impurity of the overall tree as a \emph{weighted} sum of the leaf impurities. Non-normalized Gini impurity of a leaf can take values between $\left[0, 1 - \frac{1}{|\mathcal{C}|}\right]$. An impurity value of 0 corresponds to the case where the leaf consists only of datapoints with the same class label (i.e., perfectly pure leaf). The maximum impurity value corresponds to the case where the leaf consists of an equal number of datapoints from each class. 

We chose to conduct our analyses with the two alternative loss functions described above to emphasize a conceptual difference between two options of class label assignments for datapoints within a leaf: (i) explicitly assigning a \emph{unique} class label, and (ii) assigning a probability ($p_c, \ c \in \mathcal{C}$) of belonging to each class. Misclassification error uses the former assignment option, whereas Gini impurity uses the latter. The unique label assignment in the former and the probabilistic label assignment in the latter are sometimes referred to as \emph{hard classification} and \emph{soft classification}, respectively. 

\begin{figure}
    \centering
    \includegraphics[width=16cm]{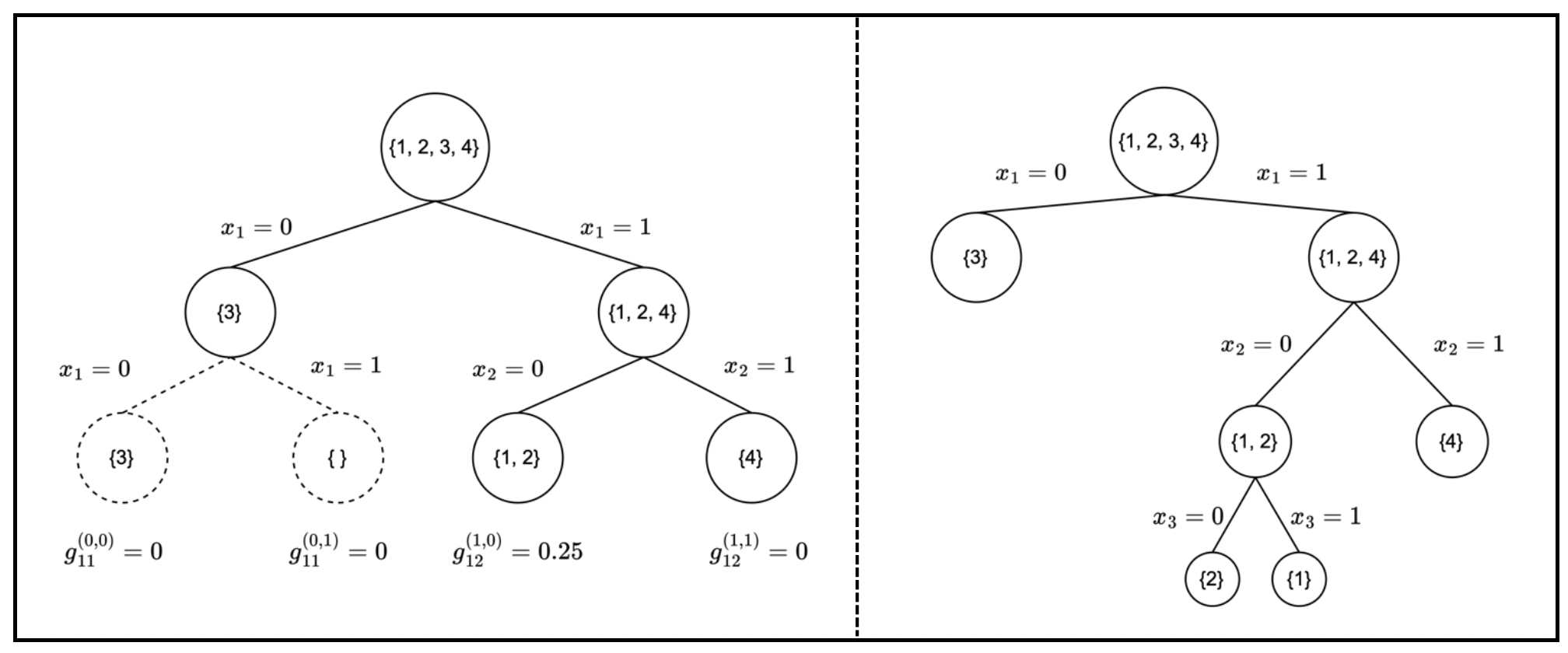}
    \caption{An optimal 2-depth tree (left panel) and 3-depth tree (right panel) when Gini impurity is used in the objective function}
    \label{fig:Gini_tree}
\end{figure}

To illustrate the effect of using different loss functions in model formulation, consider a simplistic dataset given in Table \ref{tab:example_dataset} with 4 datapoints, 3 binary features, and an outcome variable with two possible class labels A and B. At the root node, we have a baseline misclassification error of 0.25 and a baseline Gini impurity of $0.375 = \frac{4}{4} \left[1 - (\frac{1}{4})^2 - (\frac{3}{4})^2 \right]$. If we formulate the 2-depth optimal classification tree (OCT-2) with misclassification error as the objective function, we would not find any solution that improves upon the baseline misclassification, thus, the model would not produce any tree. If we formulate with Gini impurity, on the other hand, the model produces an optimal 2-depth tree shown in the left panel of Figure \ref{fig:Gini_tree} with the Gini impurity $0.25 = \frac{1}{4}\times0+\frac{0}{4}\times0+\frac{2}{4}\times\left[1-(\frac{1}{2})^2 - (\frac{1}{2})^2 \right] +\frac{1}{4}\times0$. 

\begin{table}[htbp]
  \centering
  \small
  \caption{An example dataset}
    \begin{tabular}{c|ccc|c}
    \toprule
    \multirow{2}{*}{\textbf{Datapoint}} & \multicolumn{3}{c|}{\textbf{Features}} & \multicolumn{1}{c}{\textbf{Outcome}} \\
    & \multicolumn{1}{c}{$x_1$} & \multicolumn{1}{c}{$x_2$} &
    \multicolumn{1}{c|}{$x_3$} & \multicolumn{1}{c}{$y$} \\
    \midrule
    1  & 1  & 0  & 1  & A \\ 
    2  & 1  & 0  & 0  & B \\ 
    3  & 0  & 0  & 1  & B \\ 
    4  & 1  & 1  & 1  & B \\  
    \bottomrule
    \end{tabular}%
  \label{tab:example_dataset}%
\end{table}%

The choice between misclassification or Gini impurity as the loss function could be even more critical while constructing a deeper classification tree. The right panel of Figure \ref{fig:Gini_tree} illustrates the optimal classification tree with depth 3 which has a $100\%$ accuracy ($0$ misclassification error). This tree is constructed via the proposed rolling subtree approach with Gini impurity (RST-G). However, [OCT-2] would not produce any tree at the root node if, instead, misclassification error is used. This would cause the rolling subtree algorithm with misclassification (RST-M) to get stuck on the root node and never ``roll down''. One may call this a \emph{premature termination}, since, as apparent in the right panel of Figure \ref{fig:Gini_tree}, had the algorithm looked three steps ahead instead of two, it would have actually discovered the optimal tree. This example clearly illustrates the advantage of soft over hard classification since the former allows constructing intermittent (temporary) subtrees even if they do not improve the misclassification over their root node's baseline. In our computational results section, we will highlight the counter-intuitive cases of observing higher training accuracies for RST-G than RST-M, which can be explained by premature termination. 

\section{Computational Results} \label{s:CompExps}

The proposed RST-G and RST-M algorithms are compared against several benchmark algorithms using multiple public datasets. First, we describe the experimental setup including the datasets used, their characteristics, and comparison benchmarks in subsection \ref{s:exp_setup}. Next, we address the following questions in the respective subsections:
\begin{itemize}
    \item Section 5.2: What is the resulting accuracy gain via implementing a 2-step lookahead algorithm instead of a 1-step greedy approach (e.g., CART)? How does the use of different loss functions impact this comparison?
    \item Section 5.3: Which loss function should be preferred in our proposed algorithm (e.g., RST-M vs. RST-G) for higher accuracy? What are the factors affecting this decision?  
    \item Sections 5.4, 5.5: How does our proposed algorithm perform compared to the state-of-the-art optimal and myopic methods in terms of out-of-sample accuracy as well as solution time, respectively? 
\end{itemize}

\subsection{\emph{Experimental Setup}}\label{s:exp_setup}

19 datasets, available at the UCL repository \citep{Dua2019}, are collected for the experiments. Table \ref{instance_table2} provides the list and some characteristics of these datasets. We select these particular datasets to ensure that we have good coverage across five dimensions of interest: number of data points (S(mall) if $n \leq 1000$, M(edium) if $1000 < n \leq 10,000$, or L(arge) if $n>10,000$), number of binarized features (S(mall) if $p \leq 30$, M(edium) if $30 < p \leq 100$, or L(arge) if $p>100$), number of classes (B(inary) if $|\mathcal{C}|=2$ and M(ultiple) if $|\mathcal{C}|>2$), the existence of a possible overfitting problem using CART (N(ot)E(xist) if the difference between cross-validated train and test accuracy is less than 10\% or E(xist) otherwise), and the severity of class imbalance (B(alanced) if the ratio of the minority class in the whole dataset is more than 30\%, I(mbalanced) if the ratio is between 10\% and 30\%, or H(ighly)I(mbalanced) if the ratio is less than 10\%).

\begin{table}[h]
\footnotesize
\centering
\caption{Characteristics of the datasets used in the study and their respective groupings }
\label{instance_table2}
\begin{tabular}{|l|c|c|c|c|c|c|}
\hline
\multicolumn{1}{|l|}{Dataset} & \multicolumn{1}{c|}{\begin{tabular}[c]{@{}c@{}}Number of \\ Classes\end{tabular}} & \multicolumn{1}{c|}{\begin{tabular}[c]{@{}c@{}}Number of \\ Data Points\end{tabular}} & \multicolumn{1}{c|}{\begin{tabular}[c]{@{}c@{}}Number of \\ Orig. Feat.\end{tabular}} & \multicolumn{1}{c|}{\begin{tabular}[c]{@{}c@{}}Number of \\ Bin. Feat.\end{tabular}} & \multicolumn{1}{c|}{Overfitting} & \multicolumn{1}{c|}{\begin{tabular}[c]{@{}c@{}}Class \\ Imbalance\end{tabular}} \\ \hline
adult & 2 (B) & 48842 (L) & 14 & 135 (L) & E & I \\
tae & 2 (B) & 151 (S) & 5 & 32 (M) & E & I \\
spambase & 2 (B) & 4601 (M) & 57 & 136 (L) & E & B \\
tic-tac-toe & 2 (B) & 958 (S) & 9 & 27 (S) & NE & B \\
banknote-authentication & 2 (B) & 1372 (M) & 4 & 40 (M) & NE & B \\
wine & 3 (M) & 178 (S) & 13 & 130 (L) & NE & I \\
haberman & 2 (B) & 306 (S) & 3 & 25 (S) & E & I \\
wdbc & 2 (B) & 569 (S) & 30 & 300 (L) & NE & B \\
diabetes & 2 (B) & 768 (S) & 8 & 72 (M) & E & B \\
seismic-bumps & 2 (B) & 2584 (M) & 18 & 72 (M) & E & HI \\
monks-1 & 2 (B) & 556 (S) & 6 & 15 (S) & NE & B \\
monks-2 & 2 (B) & 601 (S) & 6 & 15 (S) & NE & B \\
monks-3 & 2 (B) & 554 (S) & 6 & 15 (S) & NE & B \\
titanic & 2 (B) & 891 (S) & 8 & 178 (L) & E & B \\
kr-vs-kp & 2 (B) & 3196 (M) & 36 & 38 (M) & NE & B \\
car-evaluation & 4 (M) & 1728 (M) & 6 & 21 (S) & NE & HI \\
balance-scale & 3 (M) & 625 (S) & 4 & 20 (S) & E & HI \\
nursery & 5 (M) & 12960 (L) & 8 & 26 (S) & NE & HI \\
agaricus-lepiota & 2 (M) & 8124 (M) & 22 & 112 (L) & NE & B \\
\hline
\end{tabular}
\end{table}

Data preprocessing is applied to each dataset before we apply our algorithms which assume that all features are binary. For numerical features, we follow a two-step approach. If the number of unique values observed for a feature is less than 7, then it is treated as categorical. Otherwise, it is discretized into equal-sized groups based on sample quantiles. Next, for all these processed features and the categorical ones, we simply used one-hot encoding (i.e., introducing a dummy variable for each category). For each dataset, the total number of resulting binary features, after preprocessing, is provided in Table \ref{instance_table2}.

We evaluate the performance of our algorithms, RST-M and RST-G, and compare them against two groups of methods: myopic and optimal. For the first group, we use CART which is a commonly used benchmark. {\textit {DecisionTreeClassifier}} from scikit-learn Python library is used to implement CART with the Gini impurity criterion (CART-G). Default parameters are used except {\textit {max\_depth}} (which is changed from 2 through 8). We implement a simplified version of our RST-M algorithm that looks only one step ahead and uses misclassification as the loss function (CART-M), since {\textit {DecisionTreeClassifier}} does not handle misclassification. 

The second group of algorithms is based on global optimization methods. BinOCT \citep{verwer_learning_2019} only generates balanced binary decision trees. Thus, we use the alternative formulation of BinOCT proposed in \cite{aghaei_learning_2019}. Moreover, we evaluate FlowOCT and BendersOCT models from \cite{aghaei_strong_2022}, the codes of which are publicly available. For all these models, model building and solution times are each constrained by 30 minutes for a total of 1 hour. If the previous depth took more than 1 hour to solve or if a feasible solution could not be found in that time, we skipped solutions for deeper trees. Under these computational time limits, BendersOCT performed best in this group of methods, hence used as the benchmark optimization-based method. Details of our numerical comparison amongst this group are provided in the Online Appendix B.

Decision trees with depths ranging from 2 through 8 are created using each method. Deeper trees are sometimes needed for larger instances, which also allows testing for the scalability of an algorithm. We calculate both in-sample and out-of-sample accuracy results using $10$-fold cross-validation. To handle imbalanced datasets, we employed stratified sampling while creating the folds, which are kept the same across different methods. All models are implemented in Python programming language with a Gurobi 9.1 solver. Google Cloud Compute is used for the running environment with a virtual machine of 12 CPU and 64 GB RAM. We compare the performance of the algorithms over each of the 19 datasets using $10$-fold cross-validation at each depth, resulting in 1330 (19*10*7) problem instances in total. 

\subsection{Is it really worth looking two steps ahead instead of one?}

Computation time considerations aside, the question of \emph{whether looking two steps ahead is better than looking one step ahead when choosing the split conditions} may sound trivial as the intuitive answer would be affirmative. To be more specific, as opposed to greedily choosing a split condition based on the marginal accuracy improvement, being able to look one more step ahead has the advantage to select the split condition that gives the best accuracy when considered in conjunction with the subsequent splits. This may make sense for improving the training (in-sample) accuracy, but not necessarily for the out-of-sample accuracy due to learning pathology. 

In our first analysis, we use two tree construction algorithms that employ misclassification as the loss function: CART-M and RST-M. CART-M uses a one-step lookahead algorithm whereas RST-M uses a two-step lookahead. We compared the performance of the two algorithms against each other in terms of out-of-sample accuracy. Specifically, we calculated the number of problem instances each algorithm outperforms the other. 

\begin{figure}[h!]
    \centering
    \includegraphics[width=16cm]{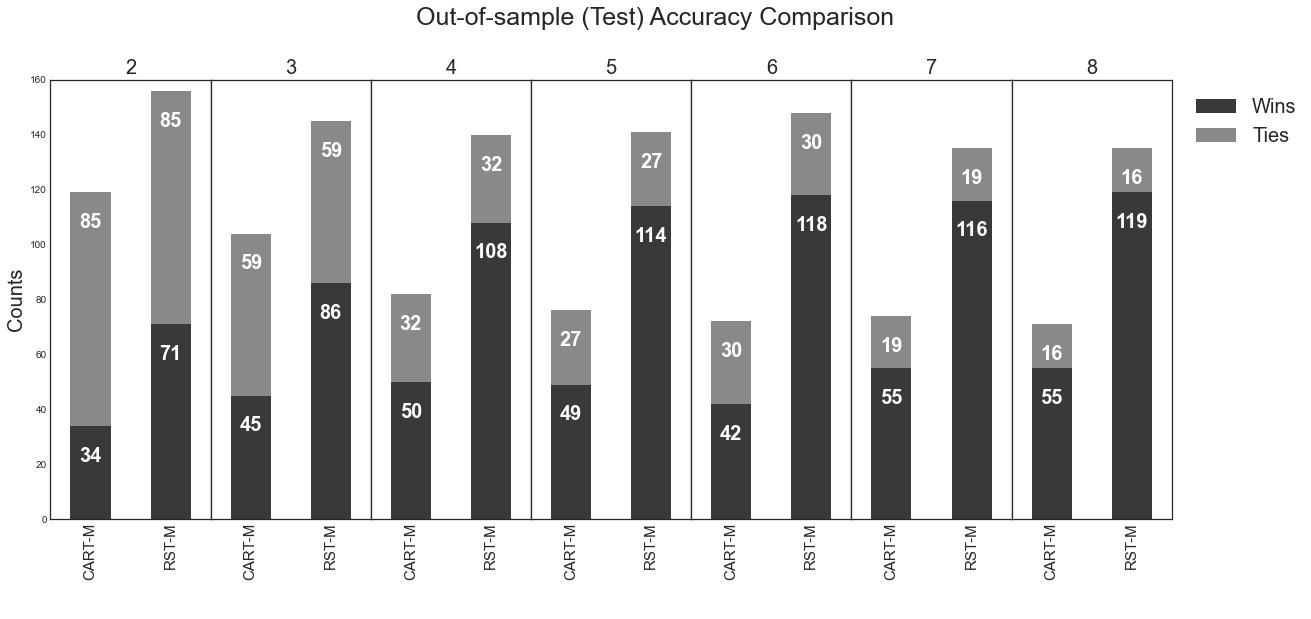}
    \caption{Comparison of CART-M vs RST-M methods in terms of out-of-sample accuracy}
    \label{fig:avsm_depth_test}
\end{figure}

Figure \ref{fig:avsm_depth_test} illustrates this comparison. Each vertical segment separated by solid lines represents the subset of 190 instances with the specific depth given at the top of the segment. Within each segment, there are two bars, one for each method. Finally, the black portions of the bars represent the number of instances the respective method strictly outperforms the other (i.e., win count), whereas the gray portion represents the number of instances where the two methods tie (i.e., tie count). Notice the tie counts are equal for the two methods within each vertical segment. It becomes evident from the win counts in Figure \ref{fig:avsm_depth_test} that RST-M consistently outperforms CART-M across all depths. Moreover, the difference between the win counts widens as the depth increases and stabilizes beyond depth 6. Especially for shallow trees (i.e., depths 2 and 3), we observe that for a significant proportion of the instances, the two methods tie in as expected. The reduction in the tie count and the widening of the win count difference indicate that the effect of looking two steps ahead instead of one manifests itself more clearly in constructing deeper trees. 

\begin{figure}[h!]
    \centering
    \includegraphics[width=16cm]{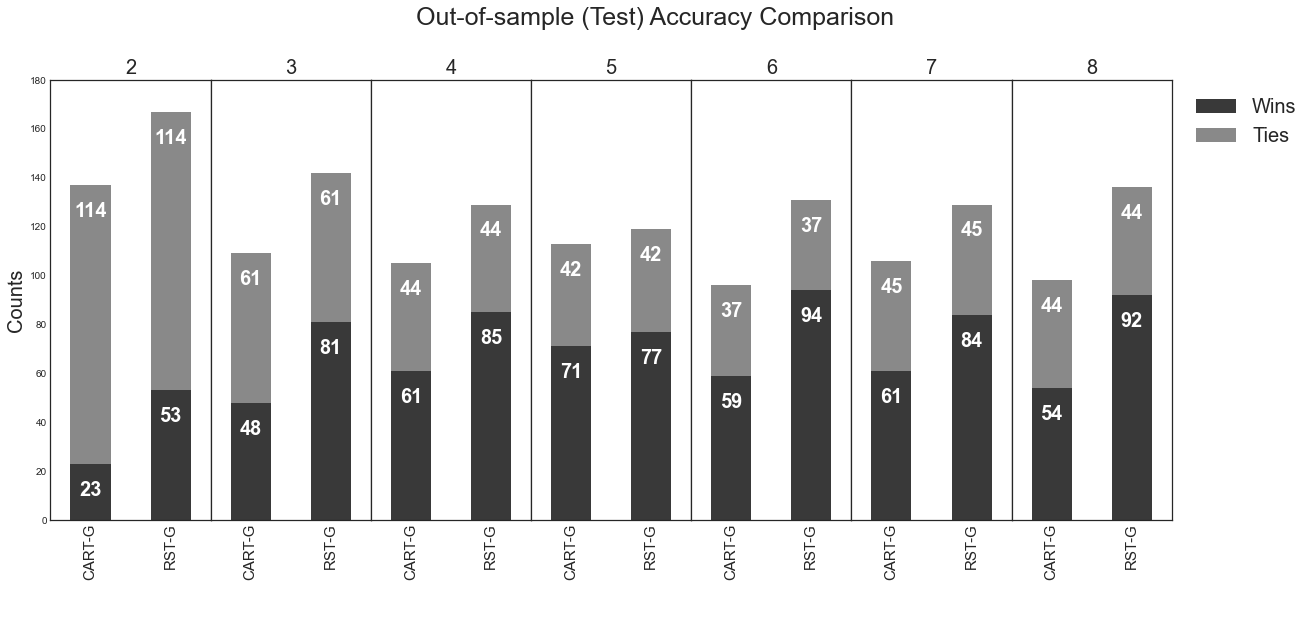}
    \caption{Comparison of CART-G vs RST-G methods in terms of out-of-sample accuracy}
    \label{fig:cart_vs_Gini_test}
\end{figure}

We, next, conduct a similar analysis using CART-G and RST-G algorithms (with the Gini impurity loss function) to see if the superior performance of the two-step lookahead is sustained when a different loss function is used. We show the comparison results in Figure \ref{fig:cart_vs_Gini_test}. A similar pattern as in Figure \ref{fig:avsm_depth_test} is observed albeit with a smaller difference in the win counts and larger tie counts. RST-G consistently outperforms CART-G with respect to the out-of-sample accuracy, confirming that the two-step lookahead is still superior to the one-step (myopic) lookahead even if we used a different loss function (i.e., Gini impurity). Therefore, we conclude that, as long as the computational costs are reasonable, it is better to look two steps ahead instead of myopically looking only one step ahead. 

\subsection{\emph{Is it better to optimize misclassification or Gini impurity?}}

Recall that we use accuracy as an evaluation metric when comparing the performance of algorithms, which is simply one minus the misclassification error. Now, we attempt to answer another seemingly trivial question: \emph{Does optimizing directly for misclassification result in a more accurate overall tree than optimizing for Gini impurity, instead?} If we use in-sample accuracy as the evaluation metric, the intuitive answer is the former, i.e., optimize directly for misclassification to get a more accurate tree. To see whether this intuitive answer is correct and quantify the result, we conduct a similar analysis as in Section 5.2. We compare the performance of the two methods (RST-M and RST-G) with respect to the in-sample accuracy for all instances across all tree depths. 

\begin{figure}
    \centering
    \includegraphics[width=16cm]{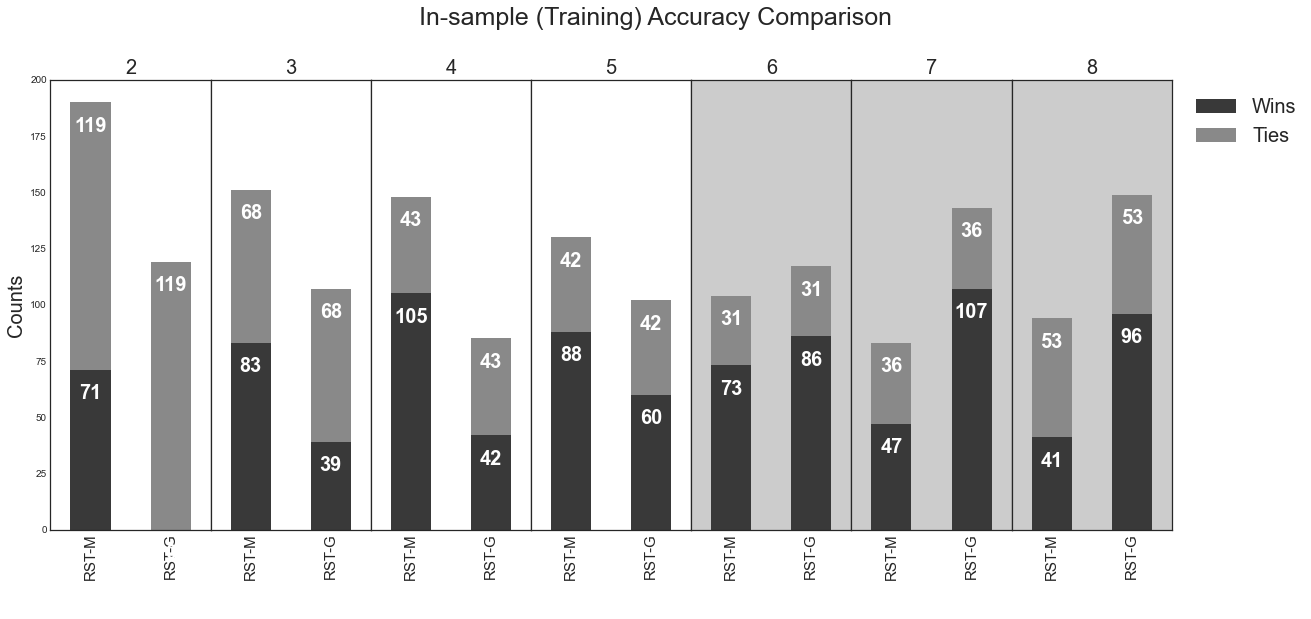}
    \caption{Comparison of RST-M and RST-G in terms of in-sample accuracy}
    \label{fig:acc_vs_Gini_training}
\end{figure}

Figure \ref{fig:acc_vs_Gini_training} shows the win and tie counts in the same fashion as in Figures \ref{fig:avsm_depth_test} and \ref{fig:cart_vs_Gini_test}. We observe that for the shallow trees (i.e., up to depth 5), RST-M outperforms RST-G in terms of in-sample accuracy, as expected. Beyond depth 5, however, RST-G's performance seems to surpass that of RST-M's. This observation implies that around depth 5, there is a \emph{switch} in the performance. In other words, if relatively shallower final trees are desired, then directly optimizing for misclassification is better when choosing the split conditions compared to optimizing for Gini impurity as a proxy. For deeper trees, however, it is more advantageous to use Gini impurity as a proxy loss function in the RST algorithm. This counter-intuitive result warrants some discussion on the characteristics of the loss functions used in tree construction. 

A possible explanation for this \emph{switching} behavior lies in the comparison between \emph{soft and hard} classification discussed in Section 4. Recall the toy example in which using Gini impurity allowed us to find an optimal tree of depth three, whereas using misclassification resulted in premature termination (i.e., no subtree of depth 2 was found that improved upon the root node's accuracy). This situation is more likely to occur in nodes where there are few datapoints and the class imbalance is very high. Since it is expected to have nodes of these characteristics as we roll down the tree, one may expect to observe this situation more frequently for deeper trees. Consequently, this could result in RST-G to start outperforming RST-M beyond a switching point which happens to be around depth 5. 

\begin{figure}[h!]
    \centering
    \includegraphics[width=16cm]{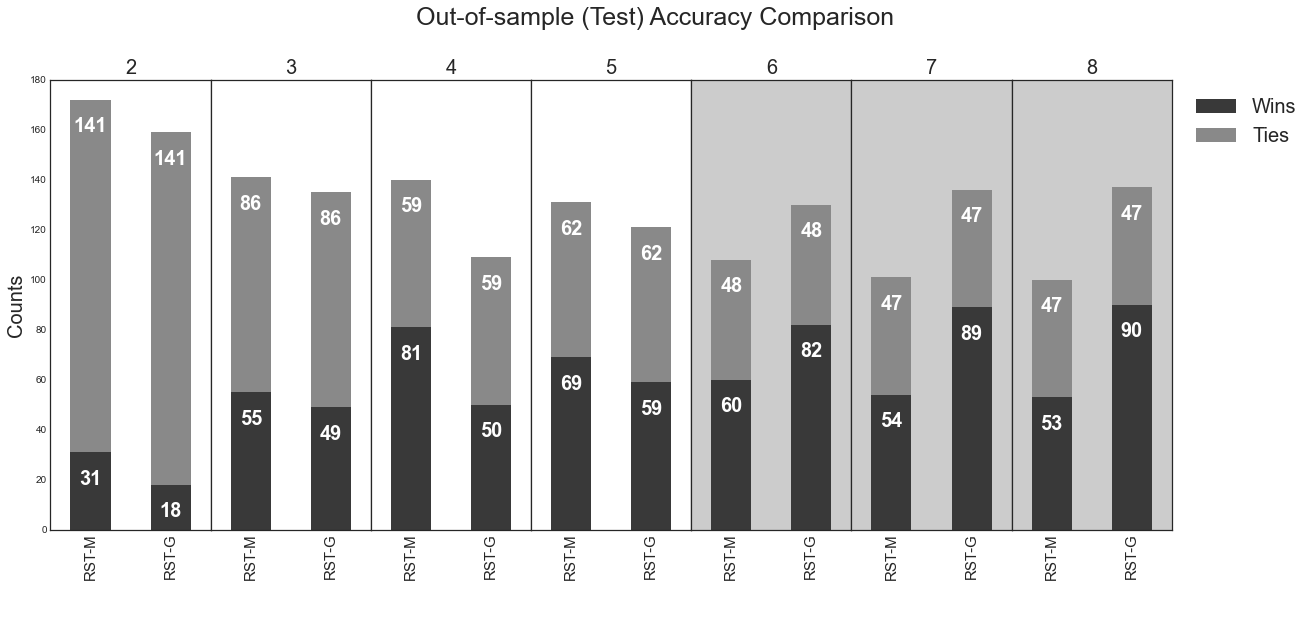}
    \caption{Comparison of RST-M and RST-G in terms of out-of-sample accuracy}
    \label{fig:acc_vs_Gini_test}
\end{figure}

Finally, we also compared the performance of the two methods using out-of-sample accuracy and observed a similar pattern, i.e., a switching behavior at around depth 5. Figure \ref{fig:acc_vs_Gini_test} illustrates this comparison. The consistent \emph{switch} beyond depth 5 motivated us to introduce a hybrid model which uses RST-M for constructing trees with depths less than or equal to five and RST-G for constructing deeper trees. Hereinafter, we refer to this hybrid approach as \emph{HybridRST}, and in the subsequent sections we compare its performance against the myopic and optimal methods. 

\subsection{\emph{Is RST really at the sweet spot between OCT and CART?}}

In this section, we try to answer the ultimate question that constitutes the main premise of the study: \emph{Is there really a sweet spot between the two extremes of decision tree construction algorithms (myopic and optimal), and if so, does RST fall in that place?} To answer this question, we conduct a three-way comparison between BendersOCT, CART-G, and HybridRST. 

\begin{figure}[h!]
    \centering
    \includegraphics[width=16cm]{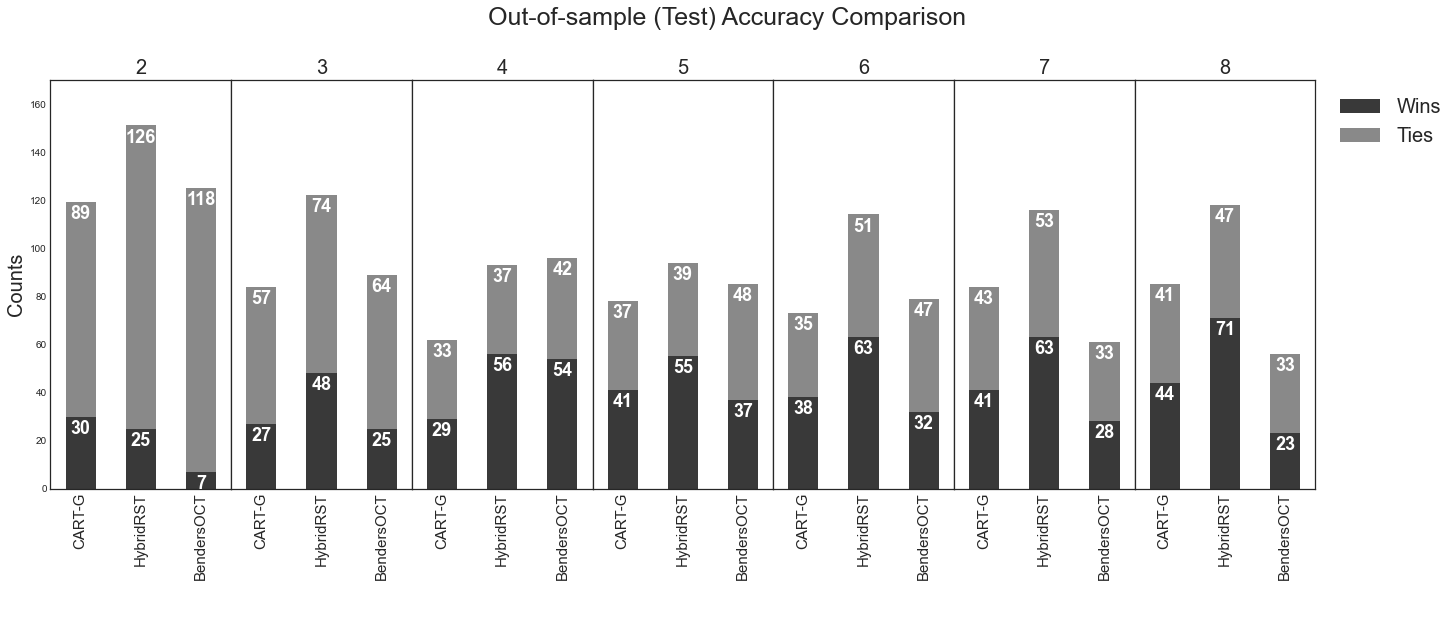}
    \caption{Three-way comparison BendersOCT, CART-G, and HybridRST in terms of out-of-sample accuracy}
    \label{fig:benders_vs_hybrid_depth_test}
\end{figure}

Figure \ref{fig:benders_vs_hybrid_depth_test} displays the win counts and the tie-for-best counts of the three methods with respect to the out-of-sample accuracy. Note that we need to take into account the tie-for-best counts in a different way than in pairwise comparison. In a pairwise comparison,  the tie counts were the same for both methods, so it was sufficient to make the comparison based solely on the win counts. In a three-way comparison, however, we need to consider the sum of win and tie-for-best counts to properly assess which method outperforms which. For example, comparing only the win counts of the first vertical block in Figure \ref{fig:benders_vs_hybrid_depth_test}, we would incorrectly conclude that CART-G is the best-performing method among the three. There are many instances where BendersOCT and HybridRST tie for best and strictly outperform CART-G. As a result, looking at the sum of the win counts and the tie-for-best counts, the HybridRST method is the best among the three. 

In general, for trees with depth less than or equal to 5, the three methods seem to be performing similarly, without a clear winner (although Hybrid performs slightly better across all depths except 4). Beyond depth 5, however, the HybridRST model starts to significantly outperform both methods. It is worth noting that this depth corresponds to the switching threshold of the HybridRST method where RST-G takes over. Moreover, we see a clear deterioration in the performance of BendersOCT beyond depth 4. This is likely because BendersOCT is rarely able to reach optimality for deeper trees. Finally, compared to CART-G, HybridRST is consistently better across all depths, with the difference between the two methods becoming more pronounced for deeper trees. Quantitatively, the total number of instances where HybridRST is the best-performing method (in terms of out-of-sample accuracy) in 808 out of 1330 instances, whereas BendersOCT and CART-G are the best in 591 and 585 instances, respectively.  

To understand the overall accuracy difference between the three methods, Table 3 provides, for each dataset included in this study, the out-of-sample accuracies of the three methods averaged across 10-folds and depths 2 through 8 (i.e., each number is the average of $10\times 7=70$ accuracy values). Bold numbers represent the best average out-of-sample accuracies among the three methods. Percentage improvements of the HybridRST over the BendersOCT (CART-G) are given in the last column. Results for specific depths are given in separate tables in the Online Appendix C which further support our conclusions.

We observe that the HybridRST method has the best average out-of-sample accuracy for 11 out of 19 datasets, whereas BendersOCT and CART-G perform the best only for 6 and 5 datasets, respectively. The datasets for which HybridRST performs the best cover all the categories listed in Table 2, whereas BendersOCT only wins for datasets with small or medium data points and features. As can be seen, HybridRST provides up to 23.6\% improvement over BendersOCT (in nursery dataset) and up to 14.4\% improvement over CART-G (in monks-1 dataset). When evaluated over 1330 problem instances, the average percentage difference between HybridRST and the best-of-three (i.e., a proxy for optimality gap) is 0.8\% (with a standard deviation of 1.2\%). These figures for CART and BendersOCT are 1.5\% (2.8\%) and 5.2\% (6.6\%), respectively. Thus we conclude that HyridRST delivers not only higher accuracies on average but does so more consistently than the alternatives.   

Although 19 datasets may not be sufficient to conclude that the RST method is unequivocally better than both the myopic and the optimal methods, looking at these results, we may, nevertheless, conclude that, in a non-negligible proportion of the cases, it falls in the sweet spot between the two extremes of a spectrum where it can improve upon both.

\begin{table}[h!]
\label{tab:test_accuracy}
\centering
\caption{Out-of-sample accuracies ($\%$) for each dataset averaged over 10-folds and depths}
\begin{tabular}{lllll}
\toprule
\bfseries Dataset&      \bfseries BendersOCT&     \bfseries CART-G&           \bfseries HybridRST & \bfseries Improvement ($\%$) \\
\midrule
adult&                  76.3&           82.1&           \bfseries 82.4 & +9.0$\%$ (+0.4$\%$) \\
agaricus-lepiota&       96.9&           99.0&           \bfseries 99.4 & +2.6$\%$ (+0.4$\%$)\\
balance-scale&          \bfseries 71.1& 70.4&           69.8  & - 1.8$\%$ (- 0.9$\%$) \\
banknote-authentication&87.8&           85.3&           \bfseries 88.6 & +0.9$\%$ (+3.9$\%$) \\
car-evaluation&         81.2&           85.0&           \bfseries 86.0 & +5.9$\%$ (+1.2$\%$) \\
diabetes&               71.2&           71.3&           \bfseries 71.7 & +0.7$\%$ (+0.6$\%$) \\
haberman&               \bfseries 68.1& \bfseries 68.1& 67.6 & - 0.7$\%$ (- 0.7$\%$) \\
kr-vs-kp&               85.0&           91.9&           \bfseries 94.7 & +11.4$\%$ (+3.1$\%$) \\
monks-1&                \bfseries 94.3& 82.4&           \bfseries 94.3 & \ 0.0$\%$ (+14.4$\%$) \\
monks-2&                64.4&           \bfseries 80.2& 76.7 &+19.1$\%$ (- 4.4$\%$) \\
monks-3&                \bfseries 98.5& 98.4&           97.0 & - 1.5$\%$ (- 1.4$\%$) \\
nursery&                71.1&           86.8&           \bfseries 87.9 & +23.6$\%$ (+1.3$\%$) \\
seismic-bumps&          \bfseries 93.1& 92.6&           92.4 & - 0.8$\%$ (- 0.2$\%$) \\
spambase&               74.4&           88.0&           \bfseries 89.2 & +19.9$\%$ (+1.4$\%$)\\
tae&                    \bfseries 83.1& 80.7&           81.6 & - 1.8$\%$ (+1.1$\%$) \\
tic-tac-toe&            77.2&           \bfseries 84.7  & 82.1 & +6.4$\%$ (- 3.1$\%$) \\
titanic&                80.7&           80.9&           \bfseries 81.0 & +0.4$\%$ (+0.1$\%$)  \\
wdbc&                   88.4&           \bfseries 89.3& 88.6 & +0.2$\%$ (- 0.8$\%$) \\
wine&                   70.4&           \bfseries 74.2& \bfseries 74.2 & +5.4$\%$ (0.0$\%$) \\
\bottomrule
\end{tabular}
\end{table}

\subsection{\emph{A note on computation times}}

In this final subsection, we compare the computation times of the three methods. Specifically, in Table 4, we present the effect of tree depth on the computation times. First, observe that CART's computation time is virtually negligible and although we observe a slight increase for deeper trees, it is always under 1 second, which shows, as expected, that (myopic) CART is unequivocally the most computationally efficient method.

\begin{table}[h!]
\caption{Comparison of computation times (in seconds) with respect to tree depth}\label{time_depth}
\begin{center}
 \begin{tabular}{|c|c|c|c|} 
 \hline
 \textbf{Depth} & \textbf{CART} & \textbf{BendersOCT} & \textbf{HybridRST}\\ [0.5ex] 
 \hline\hline
 2 & 0.006 & 357.2    &  9.712   \\
 3 & 0.007  & 1078.3  &  20.407   \\
 4 & 0.009  & 1485.9  &  31.018   \\
 5 & 0.010  & 1559.0  &  42.092   \\
 6 & 0.011  & 1604.7  &  69.759   \\
 7 & 0.012  & 1683.7  &  86.458   \\
 8 & 0.013  & 1643.5  & 107.171    \\ 
\hline
\end{tabular}
\end{center}
\end{table}

For BendersOCT, the computation times seem to be increasing nearly exponentially with depth, reaching the time limit of 1800 seconds at around depth 4. HybridRST, on the other hand, seems to scale approximately linearly with tree depth and computation times stay within a range of one to two minutes, on average. The first row of Table 4 reveals an observation that illustrates the effectiveness of the [OCT-2] formulation. Since HybridRST runs RST-M for depth 2, the (optimal) in-sample accuracies obtained by BendersOCT and HybridRST are identical for trees of depth 2. However, on average, it takes approximately 6 minutes for BendersOCT to reach the optimal solution (i.e., to prove optimality), whereas it takes less than 10 seconds for HybridRST. We conclude that these reasonable computation times qualify HybridRST as an alternative method that has the potential to improve the out-of-sample accuracy over both myopic (CART) and optimal (OCT) methods.

\section{Conclusion}\label{s:conclusion}

Despite the emergence of more complex models providing higher accuracies, decision trees continue to be popular due to their interpretability and scalability. We devise a rolling subtree algorithm with two-step lookahead to construct decision trees utilizing mixed-integer optimization. Our solution inherits the relative scalability of the myopic solutions and the ability to look multiple steps ahead of the optimal approaches. The proposed approach mitigates the learning pathology associated with the latter group of algorithms by limiting how many steps to look ahead and is flexible enough to accommodate almost any objective function during training. Even though we do not explicitly aim to find optimal trees, optimization, nevertheless, is at the center of our algorithm.    

The performance of our proposed algorithm is tested using 19 datasets. We find that our  lookahead approach is quite practical, especially for constructing deeper trees. Customizing the objective function for generating trees of different depths turns out to be quite rewarding: Optimizing only for misclassification may not be optimal, even when it is used as an out-of-sample evaluation criterion. We find that the winning recipe is a combination of both how many steps to lookahead and the employed objective function during the training phase. Our proposed hybrid algorithm follows this recipe which turns out to deliver more accurate trees on average and does so more consistently than other benchmark methods. 

Multiple avenues for future work exist to deepen our understanding of rolling subtree algorithms with few-steps lookahead. In this paper, we only consider two-step lookahead models, due to computational time considerations and recommendations in the related literature. The accuracy and computational time performances of three-step lookahead models, for which the optimization model is provided in the online appendix, could also be studied. Further speeding up our algorithm is possible via parallelization, as searching for subtrees over different branches could be done independently. Moreover, the partial enumeration stage could be made more efficient by eliminating ineffective leaf decision rules altogether via column generation. Similarly, additional constraints for split rules (e.g., fairness considerations) could be integrated into our optimization formulation which would further reduce the search space. Although we are interested in constructing trees of given depths, pruning strategies could be employed during a post-processing stage to simplify the resulting trees. 

\section*{Data Availability}

All the datasets that support the findings of this study are publicly available at the UCL repository \citep{Dua2019}.

\bibliographystyle{chicago}
\spacingset{1}
\bibliography{LatexTemplate.bib}

\begin{thebibliography}{}

\bibitem[\protect\citeauthoryear{Aghaei, Azizi, and Vayanos}{Aghaei
  et~al.}{2019}]{aghaei_learning_2019}
Aghaei, S., M.~J. Azizi, and P.~Vayanos (2019).
\newblock Learning optimal and fair decision trees for non-discriminative
  decision-making.
\newblock ~{\em 33\/}(01), 1418--1426.

\bibitem[\protect\citeauthoryear{Aghaei, Gomez, and Vayanos}{Aghaei
  et~al.}{2020}]{aghaei_learning_2020}
Aghaei, S., A.~Gomez, and P.~Vayanos (2020).
\newblock Learning optimal classification trees: Strong max-flow formulations.
\newblock {\em arXiv preprint arXiv:2002.09142\/}.

\bibitem[\protect\citeauthoryear{Aghaei, G{\'o}mez, and Vayanos}{Aghaei
  et~al.}{2021}]{aghaei_strong_2022}
Aghaei, S., A.~G{\'o}mez, and P.~Vayanos (2021).
\newblock Strong optimal classification trees.
\newblock {\em arXiv preprint arXiv:2103.15965\/}.

\bibitem[\protect\citeauthoryear{Aglin, Nijssen, and Schaus}{Aglin
  et~al.}{2020}]{aglin2020learning}
Aglin, G., S.~Nijssen, and P.~Schaus (2020).
\newblock Learning optimal decision trees using caching branch-and-bound
  search.
\newblock In {\em Proceedings of the AAAI Conference on Artificial
  Intelligence}, Volume~34, pp.\  3146--3153.

\bibitem[\protect\citeauthoryear{Angelino, Larus-Stone, Alabi, Seltzer, and
  Rudin}{Angelino et~al.}{2018}]{angelino2018learning}
Angelino, E., N.~Larus-Stone, D.~Alabi, M.~Seltzer, and C.~Rudin (2018).
\newblock Learning certifiably optimal rule lists for categorical data.
\newblock {\em Journal of Machine Learning Research\/}~{\em 18}, 1--78.

\bibitem[\protect\citeauthoryear{Bennett}{Bennett}{1992}]{bennett_decision_1992}
Bennett, K.~P. (1992).
\newblock Decision tree construction via linear programming.
\newblock Technical report, University of Wisconsin-Madison Department of
  Computer Sciences.

\bibitem[\protect\citeauthoryear{Bennett and Blue}{Bennett and
  Blue}{1996}]{Bennett96optimaldecision}
Bennett, K.~P. and J.~A. Blue (1996).
\newblock Optimal decision trees.
\newblock Technical report.

\bibitem[\protect\citeauthoryear{Bertsimas and Dunn}{Bertsimas and
  Dunn}{2017}]{bertsimas_optimal_2017}
Bertsimas, D. and J.~Dunn (2017).
\newblock Optimal classification trees.
\newblock {\em Machine Learning\/}~{\em 106\/}(7), 1039--1082.

\bibitem[\protect\citeauthoryear{Blanquero, Carrizosa, Molero-R{\'\i}o, and
  Morales}{Blanquero et~al.}{2020}]{blanquero_sparsity_2020}
Blanquero, R., E.~Carrizosa, C.~Molero-R{\'\i}o, and D.~R. Morales (2020).
\newblock Sparsity in optimal randomized classification trees.
\newblock {\em European Journal of Operational Research\/}~{\em 284\/}(1),
  255--272.

\bibitem[\protect\citeauthoryear{Blanquero, Carrizosa, Molero-R{\'\i}o, and
  Morales}{Blanquero et~al.}{2021}]{blanquero_optimal_2021}
Blanquero, R., E.~Carrizosa, C.~Molero-R{\'\i}o, and D.~R. Morales (2021).
\newblock Optimal randomized classification trees.
\newblock {\em Computers \& Operations Research\/}~{\em 132}, 105281.

\bibitem[\protect\citeauthoryear{Breiman}{Breiman}{2001}]{breiman_random_2001}
Breiman, L. (2001).
\newblock Random forests.
\newblock {\em Machine learning\/}~{\em 45\/}(1), 5--32.

\bibitem[\protect\citeauthoryear{Breiman, Friedman, Olshen, and Stone}{Breiman
  et~al.}{1984}]{gordon_classification_1984}
Breiman, L., J.~Friedman, R.~Olshen, and C.~Stone (1984).
\newblock Classification and regression trees.

\bibitem[\protect\citeauthoryear{Dantzig}{Dantzig}{1956}]{dantzig1956linear}
Dantzig, G.~B. (1956).
\newblock {\em Linear inequalities and related systems}.
\newblock Number~38. Princeton university press.

\bibitem[\protect\citeauthoryear{Demirovi{\'c}, Lukina, Hebrard, Chan, Bailey,
  Leckie, Ramamohanarao, and Stuckey}{Demirovi{\'c}
  et~al.}{2020}]{demirovic_murtree:_2020}
Demirovi{\'c}, E., A.~Lukina, E.~Hebrard, J.~Chan, J.~Bailey, C.~Leckie,
  K.~Ramamohanarao, and P.~J. Stuckey (2020).
\newblock Murtree: optimal classification trees via dynamic programming and
  search.
\newblock {\em arXiv preprint arXiv:2007.12652\/}.

\bibitem[\protect\citeauthoryear{Dua and Graff}{Dua and Graff}{2017}]{Dua2019}
Dua, D. and C.~Graff (2017).
\newblock {UCI} machine learning repository.

\bibitem[\protect\citeauthoryear{Elomaa and Malinen}{Elomaa and
  Malinen}{2005}]{elomaa2005look}
Elomaa, T. and T.~Malinen (2005).
\newblock On look-ahead and pathology in decision tree learning.
\newblock {\em Journal of Experimental \& Theoretical Artificial
  Intelligence\/}~{\em 17\/}(1-2), 19--33.

\bibitem[\protect\citeauthoryear{Esmeir and Markovitch}{Esmeir and
  Markovitch}{2004}]{esmeir2004lookahead}
Esmeir, S. and S.~Markovitch (2004).
\newblock Lookahead-based algorithms for anytime induction of decision trees.
\newblock In {\em Proceedings of the twenty-first international conference on
  Machine learning}, pp.\ ~33.

\bibitem[\protect\citeauthoryear{Friedman}{Friedman}{2001}]{friedman2001greedy}
Friedman, J.~H. (2001).
\newblock Greedy function approximation: a gradient boosting machine.
\newblock {\em Annals of statistics\/}, 1189--1232.

\bibitem[\protect\citeauthoryear{Garofalakis, Hyun, Rastogi, and
  Shim}{Garofalakis et~al.}{2003}]{garofalakis2003building}
Garofalakis, M., D.~Hyun, R.~Rastogi, and K.~Shim (2003).
\newblock Building decision trees with constraints.
\newblock {\em Data Mining and Knowledge Discovery\/}~{\em 7\/}(2), 187--214.

\bibitem[\protect\citeauthoryear{G{\"u}nl{\"u}k, Kalagnanam, Li, Menickelly,
  and Scheinberg}{G{\"u}nl{\"u}k et~al.}{2021}]{gunluk2021optimal}
G{\"u}nl{\"u}k, O., J.~Kalagnanam, M.~Li, M.~Menickelly, and K.~Scheinberg
  (2021).
\newblock Optimal decision trees for categorical data via integer programming.
\newblock {\em Journal of global optimization\/}~{\em 81\/}(1), 233--260.

\bibitem[\protect\citeauthoryear{Hoffman and Kruskal}{Hoffman and
  Kruskal}{1956}]{hoffman1956integral}
Hoffman, A. and J.~Kruskal (1956).
\newblock Integral boundary points of convex polyhedra, in “linear
  inequalities and related systems”(hw kuhn and aw tucker, eds.).
\newblock {\em Annals of Mathematical Studies\/}~(38).

\bibitem[\protect\citeauthoryear{Hu, Rudin, and Seltzer}{Hu
  et~al.}{2019}]{hu_optimal_2019}
Hu, X., C.~Rudin, and M.~Seltzer (2019).
\newblock Optimal sparse decision trees.
\newblock {\em Advances in Neural Information Processing Systems\/}~{\em 32}.

\bibitem[\protect\citeauthoryear{Last and Roizman}{Last and
  Roizman}{2013}]{last_avoiding_2013}
Last, M. and M.~Roizman (2013).
\newblock Avoiding the look-ahead pathology of decision tree learning.
\newblock {\em International journal of intelligent systems\/}~{\em 28\/}(10),
  974--987.

\bibitem[\protect\citeauthoryear{Lin, Zhong, Hu, Rudin, and Seltzer}{Lin
  et~al.}{2020}]{lin_generalized_2020}
Lin, J., C.~Zhong, D.~Hu, C.~Rudin, and M.~Seltzer (2020).
\newblock Generalized and scalable optimal sparse decision trees.
\newblock In {\em International Conference on Machine Learning}, pp.\
  6150--6160. PMLR.

\bibitem[\protect\citeauthoryear{Murthy and Salzberg}{Murthy and
  Salzberg}{1995}]{murthy_lookahead_1995}
Murthy, S. and S.~Salzberg (1995).
\newblock Lookahead and pathology in decision tree induction.
\newblock In {\em IJCAI}, pp.\  1025--1033. Citeseer.

\bibitem[\protect\citeauthoryear{Nijssen and Fromont}{Nijssen and
  Fromont}{2010}]{nijssen2010optimal}
Nijssen, S. and E.~Fromont (2010).
\newblock Optimal constraint-based decision tree induction from itemset
  lattices.
\newblock {\em Data Mining and Knowledge Discovery\/}~{\em 21\/}(1), 9--51.

\bibitem[\protect\citeauthoryear{Quinlan}{Quinlan}{1986}]{quinlan_induction_1986}
Quinlan, J.~R. (1986).
\newblock Induction of decision trees.
\newblock {\em Machine learning\/}~{\em 1\/}(1), 81--106.

\bibitem[\protect\citeauthoryear{Quinlan}{Quinlan}{1993}]{quinlan_c4.5:_1993}
Quinlan, J.~R. (1993).
\newblock {\em C4.5: {Programs} for machine learning}.
\newblock The {Morgan} {Kaufmann} series in machine learning. San Mateo,
  California: Morgan Kaufmann Publishers.

\bibitem[\protect\citeauthoryear{Rokach}{Rokach}{2016}]{rokach2016decision}
Rokach, L. (2016).
\newblock Decision forest: Twenty years of research.
\newblock {\em Information Fusion\/}~{\em 27}, 111--125.

\bibitem[\protect\citeauthoryear{Rudin}{Rudin}{2019}]{rudin2019stop}
Rudin, C. (2019).
\newblock Stop explaining black box machine learning models for high stakes
  decisions and use interpretable models instead.
\newblock {\em Nature machine intelligence\/}~{\em 1\/}(5), 206--215.

\bibitem[\protect\citeauthoryear{Sethi and Sorger}{Sethi and
  Sorger}{1991}]{sethi1991theory}
Sethi, S. and G.~Sorger (1991).
\newblock A theory of rolling horizon decision making.
\newblock {\em Annals of Operations Research\/}~{\em 29\/}(1), 387--415.

\bibitem[\protect\citeauthoryear{Verwer and Zhang}{Verwer and
  Zhang}{2017}]{Verwer_learning_2017}
Verwer, S. and Y.~Zhang (2017).
\newblock Learning decision trees with flexible constraints and objectives
  using integer optimization.
\newblock In {\em International Conference on AI and OR Techniques in
  Constraint Programming for Combinatorial Optimization Problems}, pp.\
  94--103.

\bibitem[\protect\citeauthoryear{Verwer and Zhang}{Verwer and
  Zhang}{2019}]{verwer_learning_2019}
Verwer, S. and Y.~Zhang (2019).
\newblock Learning optimal classification trees using a binary linear program
  formulation.
\newblock In {\em Proceedings of the AAAI conference on artificial
  intelligence}, Volume~33, pp.\  1625--1632.

\bibitem[\protect\citeauthoryear{Wang, Lau, and Lim}{Wang
  et~al.}{2015}]{wang2015rolling}
Wang, C., H.~C. Lau, and Y.~F. Lim (2015).
\newblock A rolling horizon auction mechanism and virtual pricing of shipping
  capacity for urban consolidation centers.
\newblock In {\em International Conference on Computational Logistics}, pp.\
  422--436. Springer.

\bibitem[\protect\citeauthoryear{Wright and Recht}{Wright and
  Recht}{2022}]{wright2022optimization}
Wright, S.~J. and B.~Recht (2022).
\newblock {\em Optimization for data analysis}.
\newblock Cambridge University Press.

\end{thebibliography}

 \def\spacingset#1{\renewcommand{\baselinestretch}%
			{#1}\small\normalsize} \spacingset{1}

\setcounter{table}{0}
\renewcommand{\thetable}{C.\arabic{table}}
\setcounter{figure}{0}
\renewcommand{\thefigure}{B.\arabic{figure}}

\newpage
\section*{Appendix}

This online appendix contains three sections. First, in Appendix A, we provide an extension of the [OCT-2] formulation as presented in the paper to the case for constructing an optimal 3-depth tree. Next, we compare the performance of three optimal approaches for decision tree construction which supports our choice of BendersOCT as the benchmark algorithm for this group. Finally, Appendix C presents in-sample accuracy results over 10 folds and 7 depths as well as detailed result tables for each of the 19 datasets and for each depth 2 through 8, both for in-sample and out-of-sample.   

\subsection*{Appendix A. 3-Depth Optimal Classification Tree Model}

We present a MIO formulation [OCT-3] for the 3-depth classification tree model in a similar way to the [OCT-2] model. Constraints (2) ensure that exactly one leaf decision rule is picked for each leaf. Constraints (3) ensure that the first 4 leaf nodes share the same split feature on the first two levels. Similarly, Constraints (4) ensure that the last 4 leaf nodes share the same split feature on the first two levels. Finally, Constraints (5) ensure that all the leaf nodes share the same split feature on the first level. Naturally, [OCT-3] has many more decision variables and constraints than [OCT-2] and is much harder to solve. Since one would need to enumerate $(p^3)$ leaf decision rules to run complete model which would require a substantial amount of pre-computation time, we didn't proceed with the three-step lookahead RST in this study. 

\begin{equation}
\begin{aligned}
\min \sum_{i \in \mathcal{F}}\sum_{j \in \mathcal{F}}\sum_{k \in \mathcal{F}}{(c_{ijk}^{(0,0,0)} + c_{ijk}^{(0,0,1)}) \cdot z_{ijk}^{(1)}} + \sum_{i \in \mathcal{F}}\sum_{j \in \mathcal{F}}\sum_{k \in \mathcal{F}}{(c_{ijk}^{(0,1,0)} + c_{ijk}^{(0,1,1)}) \cdot z_{ijk}^{(2)}} \\ 
+ \sum_{i \in \mathcal{F}}\sum_{j \in \mathcal{F}}\sum_{k \in \mathcal{F}}{(c_{ijk}^{(1,0,0)} + c_{ijk}^{(1,0,1)}) \cdot z_{ijk}^{(3)}} + \sum_{i \in \mathcal{F}}\sum_{j \in \mathcal{F}}\sum_{k \in \mathcal{F}}{(c_{ijk}^{(1,1,0)} + c_{ijk}^{(1,1,1)}) \cdot z_{ijk}^{(4)}}
\end{aligned}
\end{equation}\label{objective_three}
s.t.

\begin{equation}\label{const1_three}
    \sum_{i \in \mathcal{F}}\sum_{j \in \mathcal{F}}\sum_{k \in \mathcal{F}}{z_{ijk}^{(\ell )}} = 1, \textbf{         } \forall \ell \in \{1, 2, 3, 4 \}
\end{equation}

\begin{equation}\label{const3_three}
    \sum_{k \in \mathcal{F}}{z_{ijk}^{(1)}} = \sum_{k \in \mathcal{F}}{z_{ijk}^{(2)}}, \textbf{         } \forall i,j \in \mathcal{F}
\end{equation}

\begin{equation}\label{const4_three}
    \sum_{k \in \mathcal{F}}{z_{ijk}^{(3)}} = \sum_{k \in \mathcal{F}}{z_{ijk}^{(4)}}, \textbf{         } \forall i,j \in \mathcal{F}
\end{equation}

\begin{equation}\label{const5_three}
    \sum_{j \in \mathcal{F}}\sum_{k \in \mathcal{F}}{z_{ijk}^{(1)}} = \sum_{j \in \mathcal{F}}\sum_{k \in \mathcal{F}}{z_{ijk}^{(3)}}, \textbf{         } \forall i \in \mathcal{F}
\end{equation}

\begin{equation}\label{dvars_three}
    {z_{ijk}^{(1)}, z_{ijk}^{(2)}, z_{ijk}^{(3)}, z_{ijk}^{(4)}} \in \{0,1\}, \textbf{         } \forall i,j,k \in \mathcal{F}
\end{equation} \\

\subsection*{Appendix B. Comparison of Optimal Classification Tree Models}

There are various optimal approaches that utilize MIO to construct trees with a single iteration of model solving. We compare the performance of three highly cited methods from this family: BinOCT, FlowOCT, and BendersOCT as mentioned in Section 5.1.  

Figure 10 provides a comparison of both in-sample and out-of-sample accuracies of the three methods for all the depths and folds across the 19 datasets used in this study. BendersOCT performs the best across the three methods in both criteria. BinOCT is by far the worst performer in the in-sample accuracy criteria, which is slightly improved in the out-of-sample. FlowOCT is somewhere between BinOCT and BendersOCT, since the latter could generate solutions faster owing to the reformulation and optimality cuts used therein. Based on these observations, we use BendersOCT as a benchmark for this family of methods.     

Note that MIO formulations are quite large and for some instances, these methods do not provide any solution whatsoever within the allocated time limit. For these cases, we use the accuracy result obtained for the highest possible depth, instead.

\begin{figure}[H]
    \centering
    \includegraphics[width=14cm]{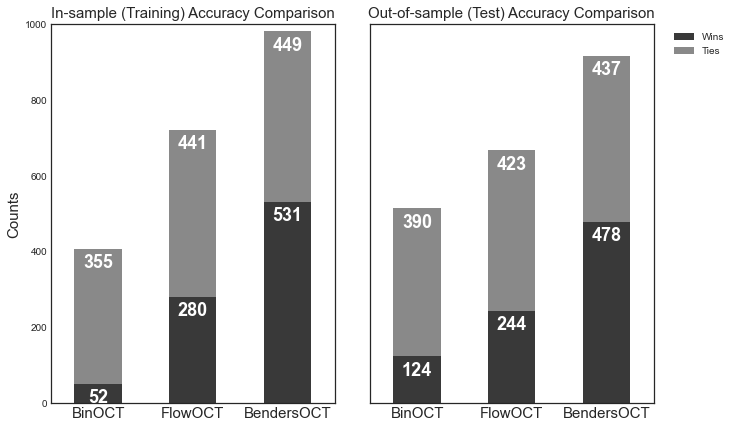}
    \caption{In-Sample and Out-of-Sample Accuracy Results for BinOCT, FlowOCT, and BendersOCT Methods}
    \label{fig:bvsf_depth}
\end{figure}

\newpage

\subsection*{Appendix C. Detailed Result Tables}

\begin{table}[h]
\label{fig:metric=training_accuracy}
\caption{In-sample accuracies (\%) for each dataset averaged over 10-folds and all depths}
\begin{tabular}{lrllllll}
\toprule
                   data &  BinOCT &        FlowOCT &     BendersOCT &           CART &          RST-M &          RST-G &         Hybrid \\
\midrule
                  adult &    76.1 &           76.6 &           76.3 &           82.3 & \bfseries 83.0 &           82.4 &           82.8 \\
       agaricus-lepiota &    72.7 &           92.9 &           96.9 &           99.0 & \bfseries 99.4 & \bfseries 99.4 & \bfseries 99.4 \\
          balance-scale &    71.1 &           78.1 &           78.5 &           78.2 &           80.9 &           79.7 & \bfseries 81.0 \\
banknote-authentication &    75.7 &           86.8 &           88.7 &           86.1 & \bfseries 89.6 &           89.5 &           89.5 \\
         car-evaluation &    75.7 &           77.4 &           82.8 &           86.2 &           87.3 &           87.1 & \bfseries 87.4 \\
               diabetes &    73.8 &           73.4 &           75.2 &           77.7 & \bfseries 81.8 &           80.1 &           80.6 \\
               haberman &    79.5 & \bfseries 82.0 &           81.9 &           79.6 &           81.9 &           80.7 &           81.7 \\
               kr-vs-kp &    78.5 &           76.7 &           85.3 &           92.3 &           94.6 &           94.8 & \bfseries 94.9 \\
                monks-1 &    91.3 & \bfseries 95.4 & \bfseries 95.4 &           85.9 &           95.0 &           94.7 &           95.0 \\
                monks-2 &    69.8 &           81.0 &           74.0 &           82.1 &           79.4 & \bfseries 83.7 &           82.0 \\
                monks-3 &    94.9 & \bfseries 98.6 & \bfseries 98.6 & \bfseries 98.6 &           96.6 & \bfseries 98.6 &           97.5 \\
                nursery &    41.7 &           82.8 &           71.0 &           86.9 &           87.8 &           88.0 & \bfseries 88.1 \\
          seismic-bumps &    93.5 &           93.8 &           93.6 &           94.4 &           94.6 & \bfseries 94.9 &           94.8 \\
               spambase &    64.4 &           77.5 &           74.8 &           89.3 & \bfseries 91.2 &           90.8 &           91.0 \\
                    tae &    92.9 & \bfseries 93.7 &           93.6 &           91.4 &           91.6 &           92.7 &           92.3 \\
            tic-tac-toe &    74.1 &           81.5 &           81.0 &           87.6 &           87.9 & \bfseries 88.5 &           88.3 \\
                titanic &    80.4 &           84.4 &           83.3 &           84.8 &           85.6 &           85.8 & \bfseries 85.9 \\
                   wdbc &    81.9 &           81.4 &           92.8 &           93.4 &           94.2 &           94.3 & \bfseries 94.4 \\
                   wine &    79.4 &           84.1 & \bfseries 84.5 &           82.5 &           83.3 &           82.3 &           83.1 \\
\bottomrule
\end{tabular}
\end{table}

\begin{table}
\label{fig:depth=2_metric=training_accuracy}
\caption{Average in-sample accuracies (\%) for each dataset for depth 2}
\begin{tabular}{llllllll}
\toprule
                   data &         BinOCT &        FlowOCT &     BendersOCT &           CART &          RST-M &          RST-G &         Hybrid \\
\midrule
                  adult &           76.1 &           76.6 &           76.8 &           79.1 & \bfseries 80.9 &           79.1 & \bfseries 80.9 \\
       agaricus-lepiota &           72.7 &           95.9 & \bfseries 96.9 &           95.4 & \bfseries 96.9 & \bfseries 96.9 & \bfseries 96.9 \\
          balance-scale & \bfseries 68.7 & \bfseries 68.7 & \bfseries 68.7 &           68.4 & \bfseries 68.7 & \bfseries 68.7 & \bfseries 68.7 \\
banknote-authentication & \bfseries 73.9 & \bfseries 73.9 & \bfseries 73.9 &           72.3 & \bfseries 73.9 & \bfseries 73.9 & \bfseries 73.9 \\
         car-evaluation & \bfseries 77.8 &           77.5 & \bfseries 77.8 & \bfseries 77.8 & \bfseries 77.8 & \bfseries 77.8 & \bfseries 77.8 \\
               diabetes & \bfseries 75.0 & \bfseries 75.0 & \bfseries 75.0 &           74.7 & \bfseries 75.0 &           74.8 & \bfseries 75.0 \\
               haberman & \bfseries 76.2 & \bfseries 76.2 & \bfseries 76.2 &           73.5 & \bfseries 76.2 &           74.5 & \bfseries 76.2 \\
               kr-vs-kp & \bfseries 86.9 & \bfseries 86.9 & \bfseries 86.9 &           77.6 & \bfseries 86.9 & \bfseries 86.9 & \bfseries 86.9 \\
                monks-1 & \bfseries 78.0 & \bfseries 78.0 & \bfseries 78.0 &           74.6 & \bfseries 78.0 & \bfseries 78.0 & \bfseries 78.0 \\
                monks-2 & \bfseries 65.7 & \bfseries 65.7 & \bfseries 65.7 & \bfseries 65.7 & \bfseries 65.7 & \bfseries 65.7 & \bfseries 65.7 \\
                monks-3 & \bfseries 96.4 & \bfseries 96.4 & \bfseries 96.4 & \bfseries 96.4 & \bfseries 96.4 & \bfseries 96.4 & \bfseries 96.4 \\
                nursery &           48.2 & \bfseries 81.2 &           76.4 &           76.4 &           76.4 &           76.4 &           76.4 \\
          seismic-bumps & \bfseries 93.5 & \bfseries 93.5 & \bfseries 93.5 &           93.4 & \bfseries 93.5 &           93.4 & \bfseries 93.5 \\
               spambase &           83.9 & \bfseries 86.1 &           82.3 &           80.1 & \bfseries 86.1 &           85.4 & \bfseries 86.1 \\
                    tae & \bfseries 84.5 & \bfseries 84.5 & \bfseries 84.5 &           83.5 & \bfseries 84.5 &           83.4 & \bfseries 84.5 \\
            tic-tac-toe & \bfseries 70.9 & \bfseries 70.9 & \bfseries 70.9 &           70.7 & \bfseries 70.9 &           70.6 & \bfseries 70.9 \\
                titanic &           81.1 &           81.1 &           81.1 &           78.9 & \bfseries 81.2 &           78.9 & \bfseries 81.2 \\
                   wdbc & \bfseries 83.1 & \bfseries 83.1 & \bfseries 83.1 &           82.9 & \bfseries 83.1 & \bfseries 83.1 & \bfseries 83.1 \\
                   wine & \bfseries 62.0 & \bfseries 62.0 & \bfseries 62.0 &           59.3 & \bfseries 62.0 &           59.6 & \bfseries 62.0 \\
\bottomrule
\end{tabular}
\end{table}

\begin{table}
\label{fig:depth=2_metric=test_accuracy}
\caption{Average out-of-sample accuracies (\%) for each dataset for depth 2}
\begin{tabular}{llllllll}
\toprule
                   data &         BinOCT &        FlowOCT &     BendersOCT &           CART &          RST-M &          RST-G &         Hybrid \\
\midrule
                  adult &           76.1 &           76.7 &           76.7 &           79.1 & \bfseries 80.9 &           79.1 & \bfseries 80.9 \\
       agaricus-lepiota &           69.4 &           94.6 & \bfseries 96.9 &           95.4 & \bfseries 96.9 & \bfseries 96.9 & \bfseries 96.9 \\
          balance-scale &           63.5 &           63.5 &           63.5 & \bfseries 66.4 &           63.5 &           63.7 &           63.5 \\
banknote-authentication & \bfseries 73.9 & \bfseries 73.9 & \bfseries 73.9 &           70.9 & \bfseries 73.9 & \bfseries 73.9 & \bfseries 73.9 \\
         car-evaluation &           77.8 & \bfseries 80.3 &           77.8 &           77.8 &           77.8 &           77.8 &           77.8 \\
               diabetes &           73.2 & \bfseries 73.5 &           73.2 & \bfseries 73.5 & \bfseries 73.5 & \bfseries 73.5 & \bfseries 73.5 \\
               haberman &           68.9 &           69.3 &           69.3 & \bfseries 73.5 &           69.0 &           70.5 &           69.0 \\
               kr-vs-kp & \bfseries 86.9 & \bfseries 86.9 & \bfseries 86.9 &           76.8 & \bfseries 86.9 & \bfseries 86.9 & \bfseries 86.9 \\
                monks-1 &           74.3 &           73.7 &           74.1 & \bfseries 74.6 &           73.4 &           73.6 &           73.4 \\
                monks-2 & \bfseries 65.7 & \bfseries 65.7 & \bfseries 65.7 & \bfseries 65.7 & \bfseries 65.7 & \bfseries 65.7 & \bfseries 65.7 \\
                monks-3 & \bfseries 96.4 & \bfseries 96.4 & \bfseries 96.4 & \bfseries 96.4 & \bfseries 96.4 & \bfseries 96.4 & \bfseries 96.4 \\
                nursery &           48.3 & \bfseries 81.0 &           76.3 &           76.3 &           76.3 &           76.3 &           76.3 \\
          seismic-bumps &           93.1 &           93.1 &           93.1 & \bfseries 93.4 &           93.2 & \bfseries 93.4 &           93.2 \\
               spambase &           83.6 & \bfseries 85.8 &           81.7 &           79.8 &           85.6 &           84.2 &           85.6 \\
                    tae &           79.4 &           79.4 &           76.8 &           79.5 &           80.1 & \bfseries 80.8 &           80.1 \\
            tic-tac-toe &           66.7 &           66.7 & \bfseries 67.0 &           66.6 &           66.8 &           66.1 &           66.8 \\
                titanic &           81.1 &           81.1 &           81.1 &           77.2 & \bfseries 81.2 &           77.2 & \bfseries 81.2 \\
                   wdbc &           79.8 &           78.7 &           79.4 & \bfseries 80.7 &           78.8 &           78.6 &           78.8 \\
                   wine &           48.9 &           47.8 &           48.3 &           50.0 &           49.5 & \bfseries 55.1 &           49.5 \\
\bottomrule
\end{tabular}
\end{table}

\begin{table}
\label{fig:depth=3_metric=training_accuracy}
\caption{Average in-sample accuracies (\%) for each dataset for depth 3}
\begin{tabular}{llllllll}
\toprule
                   data &         BinOCT &        FlowOCT &     BendersOCT &           CART &          RST-M &          RST-G &         Hybrid \\
\midrule
                  adult &           76.1 &           76.6 &           76.3 &           82.1 & \bfseries 82.2 &           82.1 & \bfseries 82.2 \\
       agaricus-lepiota &           72.7 &           92.4 &           96.8 &           98.5 & \bfseries 99.1 &           99.0 & \bfseries 99.1 \\
          balance-scale &           74.4 & \bfseries 74.6 & \bfseries 74.6 &           69.9 &           74.0 &           70.5 &           74.0 \\
banknote-authentication & \bfseries 81.7 & \bfseries 81.7 & \bfseries 81.7 &           78.8 &           81.4 &           81.4 &           81.4 \\
         car-evaluation &           80.4 &           80.8 & \bfseries 81.2 &           80.7 &           81.1 &           81.1 &           81.1 \\
               diabetes &           76.5 &           76.4 &           76.6 &           75.3 & \bfseries 76.8 &           76.1 & \bfseries 76.8 \\
               haberman & \bfseries 78.3 &           78.2 & \bfseries 78.3 &           74.9 &           78.2 &           76.0 &           78.2 \\
               kr-vs-kp &           84.2 &           74.0 &           93.4 &           90.4 & \bfseries 93.8 & \bfseries 93.8 & \bfseries 93.8 \\
                monks-1 & \bfseries 89.8 & \bfseries 89.8 & \bfseries 89.8 &           81.6 &           89.3 &           88.1 &           89.3 \\
                monks-2 & \bfseries 67.7 &           67.6 &           67.6 &           66.2 &           66.0 &           66.5 &           66.0 \\
                monks-3 & \bfseries 98.9 & \bfseries 98.9 & \bfseries 98.9 & \bfseries 98.9 &           96.4 & \bfseries 98.9 &           96.4 \\
                nursery &           51.0 &           81.2 &           78.0 & \bfseries 82.5 & \bfseries 82.5 &           82.4 & \bfseries 82.5 \\
          seismic-bumps &           93.6 &           93.6 &           93.6 &           93.5 & \bfseries 93.7 & \bfseries 93.7 & \bfseries 93.7 \\
               spambase &           61.4 &           80.5 &           80.2 &           87.4 & \bfseries 89.1 & \bfseries 89.1 & \bfseries 89.1 \\
                    tae & \bfseries 89.0 & \bfseries 89.0 & \bfseries 89.0 &           86.5 &           87.7 &           88.4 &           87.7 \\
            tic-tac-toe &           77.2 & \bfseries 77.5 &           77.3 &           75.5 &           77.4 &           76.2 &           77.4 \\
                titanic &           83.4 &           83.4 & \bfseries 83.6 &           82.3 &           82.2 &           83.4 &           82.2 \\
                   wdbc &           92.1 & \bfseries 92.2 & \bfseries 92.2 &           90.9 &           91.0 &           90.9 &           91.0 \\
                   wine & \bfseries 71.6 & \bfseries 71.6 & \bfseries 71.6 &           69.8 &           71.5 &           70.8 &           71.5 \\
\bottomrule
\end{tabular}
\end{table}

\begin{table}
\label{fig:depth=3_metric=test_accuracy}
\caption{Average out-of-sample accuracies (\%) for each dataset for depth 3}
\begin{tabular}{llllllll}
\toprule
                   data &         BinOCT &        FlowOCT &     BendersOCT &           CART &          RST-M &          RST-G &         Hybrid \\
\midrule
                  adult &           76.1 &           76.7 &           76.3 &           82.0 & \bfseries 82.1 &           82.0 & \bfseries 82.1 \\
       agaricus-lepiota &           69.4 &           91.3 &           96.7 &           98.5 & \bfseries 99.1 &           98.8 & \bfseries 99.1 \\
          balance-scale &           66.4 & \bfseries 67.6 & \bfseries 67.6 &           65.8 &           67.0 &           62.4 &           67.0 \\
banknote-authentication &           80.8 &           80.6 &           80.5 &           78.1 & \bfseries 80.9 & \bfseries 80.9 & \bfseries 80.9 \\
         car-evaluation &           79.0 &           78.0 &           79.4 &           79.0 &           80.4 & \bfseries 81.2 &           80.4 \\
               diabetes &           72.2 & \bfseries 73.2 &           72.0 &           72.6 &           72.6 &           71.7 &           72.6 \\
               haberman & \bfseries 68.9 &           68.3 &           67.6 &           68.0 & \bfseries 68.9 &           66.7 & \bfseries 68.9 \\
               kr-vs-kp &           85.1 &           72.8 &           93.5 &           90.4 & \bfseries 93.8 & \bfseries 93.8 & \bfseries 93.8 \\
                monks-1 &           86.9 &           86.7 &           86.3 &           79.8 & \bfseries 88.3 &           87.6 & \bfseries 88.3 \\
                monks-2 &           57.1 &           56.7 &           57.2 &           63.1 &           62.7 & \bfseries 63.9 &           62.7 \\
                monks-3 &           98.9 &           98.9 & \bfseries 99.1 &           98.9 &           96.4 &           98.9 &           96.4 \\
                nursery &           50.9 &           81.0 &           78.0 & \bfseries 82.5 & \bfseries 82.5 &           81.8 & \bfseries 82.5 \\
          seismic-bumps &           93.0 & \bfseries 93.1 &           93.0 &           93.0 &           92.7 &           92.8 &           92.7 \\
               spambase &           61.4 &           80.0 &           79.0 &           87.4 &           88.6 & \bfseries 88.9 &           88.6 \\
                    tae & \bfseries 82.2 &           82.1 & \bfseries 82.2 &           77.5 &           78.8 &           82.1 &           78.8 \\
            tic-tac-toe &           72.1 &           72.8 & \bfseries 73.6 &           72.5 &           72.8 &           72.0 &           72.8 \\
                titanic &           82.9 &           82.4 & \bfseries 83.2 &           81.9 &           81.0 & \bfseries 83.2 &           81.0 \\
                   wdbc &           89.8 & \bfseries 90.2 &           89.6 &           88.8 &           86.8 &           87.2 &           86.8 \\
                   wine &           59.6 &           60.2 &           58.5 &           57.8 & \bfseries 62.4 &           61.2 & \bfseries 62.4 \\
\bottomrule
\end{tabular}
\end{table}

\begin{table}
\label{fig:depth=4_metric=training_accuracy}
\caption{Average in-sample accuracies (\%) for each dataset for depth 4}
\begin{tabular}{llllllll}
\toprule
                   data &          BinOCT &         FlowOCT &      BendersOCT &           CART &           RST-M &           RST-G &          Hybrid \\
\midrule
                  adult &            76.1 &            76.6 &            76.3 &           82.2 &  \bfseries 82.8 &            82.3 &  \bfseries 82.8 \\
       agaricus-lepiota &            72.7 &            92.4 &            98.4 &           99.4 & \bfseries 100.0 & \bfseries 100.0 & \bfseries 100.0 \\
          balance-scale &            77.3 &            77.9 &  \bfseries 78.0 &           71.3 &            77.5 &            73.7 &            77.5 \\
banknote-authentication &            86.0 &  \bfseries 89.5 &  \bfseries 89.5 &           84.9 &            88.8 &            88.8 &            88.8 \\
         car-evaluation &            81.4 &  \bfseries 84.3 &            84.0 &           81.7 &            83.2 &            82.8 &            83.2 \\
               diabetes &            77.4 &            76.5 &            76.5 &           75.9 &  \bfseries 79.0 &            77.6 &  \bfseries 79.0 \\
               haberman &  \bfseries 80.2 &            80.0 &            80.1 &           77.8 &  \bfseries 80.2 &            78.3 &  \bfseries 80.2 \\
               kr-vs-kp &            74.1 &            79.8 &  \bfseries 94.6 &           94.1 &            94.4 &            94.0 &            94.4 \\
                monks-1 & \bfseries 100.0 & \bfseries 100.0 & \bfseries 100.0 &           84.6 &            97.6 &            96.6 &            97.6 \\
                monks-2 &            70.1 &            70.1 &  \bfseries 70.2 &           68.5 &            68.7 &            69.4 &            68.7 \\
                monks-3 &  \bfseries 98.9 &  \bfseries 98.9 &  \bfseries 98.9 & \bfseries 98.9 &            96.4 &  \bfseries 98.9 &            96.4 \\
                nursery &            38.6 &            83.4 &            70.0 &           85.3 &  \bfseries 87.8 &            87.2 &  \bfseries 87.8 \\
          seismic-bumps &            93.6 &            93.8 &            93.7 &           93.8 &            94.0 &  \bfseries 94.2 &            94.0 \\
               spambase &            62.2 &            81.7 &            72.6 &           89.7 &  \bfseries 90.4 &            90.1 &  \bfseries 90.4 \\
                    tae &            93.2 &            93.5 &  \bfseries 93.6 &           88.8 &            90.7 &            92.3 &            90.7 \\
            tic-tac-toe &            83.0 &            83.7 &            83.3 &           83.9 &  \bfseries 84.9 &            84.8 &  \bfseries 84.9 \\
                titanic &            84.1 &            84.3 &  \bfseries 84.7 &           84.3 &            84.3 &            84.4 &            84.3 \\
                   wdbc &            92.8 &            95.2 &  \bfseries 96.9 &           93.0 &            94.0 &            93.7 &            94.0 \\
                   wine &            80.0 &  \bfseries 80.1 &  \bfseries 80.1 &           78.5 &            79.3 &            77.8 &            79.3 \\
\bottomrule
\end{tabular}
\end{table}

\begin{table}
\label{fig:depth=4_metric=test_accuracy}
\caption{Average out-of-sample accuracies (\%) for each dataset for depth 4}
\begin{tabular}{llllllll}
\toprule
                   data &          BinOCT &         FlowOCT &      BendersOCT &           CART &           RST-M &           RST-G &          Hybrid \\
\midrule
                  adult &            76.1 &            76.7 &            76.3 &           82.2 &  \bfseries 82.7 &            82.2 &  \bfseries 82.7 \\
       agaricus-lepiota &            69.4 &            91.3 &            98.4 &           99.2 & \bfseries 100.0 & \bfseries 100.0 & \bfseries 100.0 \\
          balance-scale &            69.9 &  \bfseries 70.7 &            70.2 &           65.1 &            68.2 &            64.2 &            68.2 \\
banknote-authentication &            85.7 &  \bfseries 89.3 &            89.2 &           84.8 &            88.5 &            88.5 &            88.5 \\
         car-evaluation &            81.2 &  \bfseries 84.4 &            81.6 &           79.6 &            81.1 &            81.3 &            81.1 \\
               diabetes &            70.3 &            73.3 &  \bfseries 73.5 &           72.1 &            72.4 &            70.8 &            72.4 \\
               haberman &            68.6 &  \bfseries 70.9 &            68.3 &           69.0 &            68.6 &            67.0 &            68.6 \\
               kr-vs-kp &            74.0 &            79.0 &  \bfseries 94.4 &           94.1 &            94.1 &            93.2 &            94.1 \\
                monks-1 & \bfseries 100.0 & \bfseries 100.0 & \bfseries 100.0 &           82.0 &            98.2 &            96.6 &            98.2 \\
                monks-2 &            54.3 &            58.6 &            52.7 & \bfseries 63.2 &            54.9 &            58.9 &            54.9 \\
                monks-3 &            98.9 &            98.9 &  \bfseries 99.1 &           98.9 &            96.4 &            98.9 &            96.4 \\
                nursery &            41.0 &            84.9 &            69.9 &           85.3 &  \bfseries 87.8 &            87.2 &  \bfseries 87.8 \\
          seismic-bumps &  \bfseries 93.3 &            92.9 &            92.9 &           92.9 &            92.7 &            92.8 &            92.7 \\
               spambase &            62.2 &            80.1 &            72.3 & \bfseries 89.5 &            89.4 &            89.0 &            89.4 \\
                    tae &            80.8 &            82.1 &  \bfseries 85.4 &           76.1 &            80.1 &            84.7 &            80.1 \\
            tic-tac-toe &            78.6 &            79.2 &            77.7 & \bfseries 82.1 &            81.5 &            81.1 &            81.5 \\
                titanic &            81.2 &            82.3 &            81.2 &           82.6 &  \bfseries 82.7 &            81.5 &  \bfseries 82.7 \\
                   wdbc &            90.5 &            92.6 &  \bfseries 93.0 &           90.0 &            90.9 &            90.4 &            90.9 \\
                   wine &  \bfseries 72.5 &            68.7 &            69.8 &           68.0 &            68.0 &  \bfseries 72.5 &            68.0 \\
\bottomrule
\end{tabular}
\end{table}

\begin{table}
\label{fig:depth=5_metric=training_accuracy}
\caption{Average in-sample accuracies (\%) for each dataset for depth 5}
\begin{tabular}{llllllll}
\toprule
                   data &          BinOCT &         FlowOCT &      BendersOCT &           CART &           RST-M &           RST-G &          Hybrid \\
\midrule
                  adult &            76.1 &            76.6 &            76.3 &           82.8 &  \bfseries 83.2 &            82.6 &  \bfseries 83.2 \\
       agaricus-lepiota &            72.7 &            92.4 &            99.8 &           99.9 & \bfseries 100.0 & \bfseries 100.0 & \bfseries 100.0 \\
          balance-scale &            78.6 &            81.0 &            81.2 &           78.7 &  \bfseries 81.4 &            80.0 &  \bfseries 81.4 \\
banknote-authentication &            72.1 &  \bfseries 93.6 &  \bfseries 93.6 &           88.2 &            92.8 &            92.8 &            92.8 \\
         car-evaluation &            75.7 &            84.6 &            86.1 &           86.8 &  \bfseries 87.1 &            85.4 &  \bfseries 87.1 \\
               diabetes &            74.9 &            79.5 &            78.6 &           77.2 &  \bfseries 81.5 &            79.8 &  \bfseries 81.5 \\
               haberman &            82.2 &            82.3 &  \bfseries 82.4 &           79.8 &            82.1 &            80.6 &            82.1 \\
               kr-vs-kp &            76.1 &            85.4 &            91.9 &           94.1 &  \bfseries 95.2 &            94.7 &  \bfseries 95.2 \\
                monks-1 & \bfseries 100.0 & \bfseries 100.0 & \bfseries 100.0 &           86.3 & \bfseries 100.0 & \bfseries 100.0 & \bfseries 100.0 \\
                monks-2 &            76.7 &            82.2 &            79.3 &           78.8 &            73.4 &  \bfseries 84.1 &            73.4 \\
                monks-3 &  \bfseries 98.9 &  \bfseries 98.9 &  \bfseries 98.9 & \bfseries 98.9 &            96.4 &  \bfseries 98.9 &            96.4 \\
                nursery &            38.6 &            83.4 &            68.9 &           87.6 &  \bfseries 89.1 &            89.0 &  \bfseries 89.1 \\
          seismic-bumps &            93.5 &            93.9 &            93.7 &           94.1 &            94.4 &  \bfseries 94.8 &            94.4 \\
               spambase &            60.9 &            72.9 &            71.5 &           90.3 &  \bfseries 91.7 &            90.8 &  \bfseries 91.7 \\
                    tae &            96.1 &  \bfseries 96.2 &  \bfseries 96.2 &           92.7 &            92.7 &            94.3 &            92.7 \\
            tic-tac-toe &            77.2 &            86.1 &            86.2 &           91.1 &            89.6 &  \bfseries 92.3 &            89.6 \\
                titanic &            84.1 &            85.4 &            85.0 &           85.2 &            85.9 &  \bfseries 86.1 &            85.9 \\
                   wdbc &            76.9 &            93.6 &            96.0 &           94.8 &            96.1 &  \bfseries 96.3 &            96.1 \\
                   wine &            85.6 &  \bfseries 87.8 &  \bfseries 87.8 &           85.5 &            85.8 &            85.3 &            85.8 \\
\bottomrule
\end{tabular}
\end{table}

\begin{table}
\label{fig:depth=5_metric=test_accuracy}
\caption{Average out-of-sample accuracies (\%) for each dataset for depth 5}
\begin{tabular}{llllllll}
\toprule
                   data &          BinOCT &         FlowOCT &      BendersOCT &           CART &           RST-M &           RST-G &          Hybrid \\
\midrule
                  adult &            76.1 &            76.7 &            76.3 &           82.7 &  \bfseries 83.0 &            82.2 &  \bfseries 83.0 \\
       agaricus-lepiota &            69.4 &            91.3 &            99.8 &           99.9 & \bfseries 100.0 & \bfseries 100.0 & \bfseries 100.0 \\
          balance-scale &            71.5 &            72.6 &  \bfseries 73.8 &           70.9 &            70.6 &            70.1 &            70.6 \\
banknote-authentication &            72.6 &            92.3 &            92.1 &           87.8 &  \bfseries 92.6 &  \bfseries 92.6 &  \bfseries 92.6 \\
         car-evaluation &            76.1 &            82.7 &            83.5 & \bfseries 86.0 &            85.7 &            84.3 &            85.7 \\
               diabetes &            72.5 &            71.6 &  \bfseries 73.2 &           71.3 &            72.6 &            70.0 &            72.6 \\
               haberman &            67.6 &  \bfseries 68.3 &            67.6 &           67.0 &  \bfseries 68.3 &            67.0 &  \bfseries 68.3 \\
               kr-vs-kp &            75.8 &            84.8 &            91.1 &           94.1 &  \bfseries 95.2 &            94.1 &  \bfseries 95.2 \\
                monks-1 & \bfseries 100.0 & \bfseries 100.0 & \bfseries 100.0 &           82.3 & \bfseries 100.0 & \bfseries 100.0 & \bfseries 100.0 \\
                monks-2 &            68.7 &  \bfseries 78.0 &            71.4 &           75.2 &            53.4 &            75.9 &            53.4 \\
                monks-3 &            98.9 &            98.9 &  \bfseries 99.1 &           98.9 &            96.4 &            98.9 &            96.4 \\
                nursery &            41.0 &            84.9 &            68.8 &           87.6 &  \bfseries 88.9 &            88.8 &  \bfseries 88.9 \\
          seismic-bumps &  \bfseries 93.3 &            92.9 &            93.1 &           93.0 &            92.4 &            92.2 &            92.4 \\
               spambase &            62.2 &            73.4 &            71.5 &           89.4 &  \bfseries 90.0 &            89.5 &  \bfseries 90.0 \\
                    tae &            82.8 &            81.4 &            84.7 &           82.1 &            80.8 &  \bfseries 86.0 &            80.8 \\
            tic-tac-toe &            74.6 &            79.8 &            82.2 & \bfseries 90.6 &            82.1 &            89.5 &            82.1 \\
                titanic &            81.6 &            80.8 &            81.9 & \bfseries 82.5 &            82.2 &            80.6 &            82.2 \\
                   wdbc &            73.4 &            90.9 &            90.0 & \bfseries 91.8 &            91.0 &            90.0 &            91.0 \\
                   wine &            79.3 &  \bfseries 79.9 &            79.3 &           78.7 &            74.3 &            79.8 &            74.3 \\
\bottomrule
\end{tabular}
\end{table}

\begin{table}
\label{fig:depth=6_metric=training_accuracy}
\caption{Average in-sample accuracies (\%) for each dataset for depth 6}
\begin{tabular}{lrllllll}
\toprule
                   data &  BinOCT &         FlowOCT &      BendersOCT &            CART &           RST-M &           RST-G &          Hybrid \\
\midrule
                  adult &    76.1 &            76.6 &            76.3 &            83.1 &  \bfseries 83.5 &            83.1 &            83.1 \\
       agaricus-lepiota &    72.7 &            92.4 &            91.3 & \bfseries 100.0 & \bfseries 100.0 & \bfseries 100.0 & \bfseries 100.0 \\
          balance-scale &    67.1 &            82.4 &            83.7 &            83.5 &  \bfseries 84.9 &            84.8 &            84.8 \\
banknote-authentication &    72.1 &            92.2 &            95.0 &            89.7 &  \bfseries 95.3 &            94.5 &            94.5 \\
         car-evaluation &    71.4 &            74.8 &            85.0 &            89.0 &  \bfseries 91.1 &            90.6 &            90.6 \\
               diabetes &    72.3 &            77.4 &            74.8 &            78.5 &  \bfseries 84.2 &            81.8 &            81.8 \\
               haberman &    83.0 &            84.4 &  \bfseries 84.8 &            82.2 &            84.0 &            83.4 &            83.4 \\
               kr-vs-kp &    76.1 &            84.8 &            80.3 &            94.3 &            96.0 &  \bfseries 96.4 &  \bfseries 96.4 \\
                monks-1 &    92.6 & \bfseries 100.0 & \bfseries 100.0 &            87.3 & \bfseries 100.0 & \bfseries 100.0 & \bfseries 100.0 \\
                monks-2 &    71.3 &            95.2 &            86.3 &            96.5 &            84.7 & \bfseries 100.0 & \bfseries 100.0 \\
                monks-3 &    95.8 &  \bfseries 98.9 &  \bfseries 98.9 &  \bfseries 98.9 &            96.4 &  \bfseries 98.9 &  \bfseries 98.9 \\
                nursery &    38.6 &            83.4 &            59.2 &            89.7 &            91.1 &  \bfseries 91.7 &  \bfseries 91.7 \\
          seismic-bumps &    93.5 &            93.8 &            93.6 &            94.7 &            95.0 &  \bfseries 95.5 &  \bfseries 95.5 \\
               spambase &    60.9 &            73.8 &            71.9 &            91.4 &  \bfseries 92.8 &            91.9 &            91.9 \\
                    tae &    97.0 &  \bfseries 97.5 &            97.1 &            95.0 &            93.8 &            95.9 &            95.9 \\
            tic-tac-toe &    71.1 &            84.2 &            87.5 &            95.0 &            94.8 &  \bfseries 96.8 &  \bfseries 96.8 \\
                titanic &    76.8 &            85.4 &            84.9 &            86.3 &            87.2 &  \bfseries 87.5 &  \bfseries 87.5 \\
                   wdbc &    76.6 &            94.4 &            94.7 &            96.5 &            97.7 &  \bfseries 97.9 &  \bfseries 97.9 \\
                   wine &    87.2 &            93.5 &  \bfseries 93.8 &            91.5 &            91.4 &            90.9 &            90.9 \\
\bottomrule
\end{tabular}
\end{table}

\begin{table}
\label{fig:depth=6_metric=test_accuracy}
\caption{Average out-of-sample accuracies (\%) for each dataset for depth 6}
\begin{tabular}{llllllll}
\toprule
                   data &         BinOCT &         FlowOCT &      BendersOCT &            CART &           RST-M &           RST-G &          Hybrid \\
\midrule
                  adult &           76.1 &            76.7 &            76.3 &            82.8 &  \bfseries 83.0 &            82.8 &            82.8 \\
       agaricus-lepiota &           69.4 &            91.3 &            91.3 & \bfseries 100.0 & \bfseries 100.0 & \bfseries 100.0 & \bfseries 100.0 \\
          balance-scale &           63.9 &            73.8 &  \bfseries 75.7 &            73.0 &            73.1 &            73.1 &            73.1 \\
banknote-authentication &           72.6 &            91.3 &            94.0 &            89.1 &  \bfseries 95.0 &            93.1 &            93.1 \\
         car-evaluation &           71.2 &            72.3 &            82.9 &            86.6 &  \bfseries 88.9 &            88.3 &            88.3 \\
               diabetes &           71.3 &  \bfseries 73.2 &            70.4 &            70.3 &            71.6 &            70.3 &            70.3 \\
               haberman &           66.6 &  \bfseries 69.6 &            67.6 &            66.4 &            67.6 &            67.3 &            67.3 \\
               kr-vs-kp &           75.8 &            85.2 &            79.8 &            93.6 &            96.1 &  \bfseries 96.4 &  \bfseries 96.4 \\
                monks-1 &           92.1 & \bfseries 100.0 & \bfseries 100.0 &            80.9 & \bfseries 100.0 & \bfseries 100.0 & \bfseries 100.0 \\
                monks-2 &           67.4 &            92.3 &            76.6 &            95.0 &            75.4 & \bfseries 100.0 & \bfseries 100.0 \\
                monks-3 &           95.4 &            98.7 &            98.7 &  \bfseries 98.9 &            96.0 &            98.7 &            98.7 \\
                nursery &           41.0 &            84.9 &            59.2 &            89.6 &            90.9 &  \bfseries 91.3 &  \bfseries 91.3 \\
          seismic-bumps & \bfseries 93.3 &            92.9 &            93.2 &            92.4 &            92.1 &            92.0 &            92.0 \\
               spambase &           62.2 &            73.8 &            71.8 &            89.8 &  \bfseries 90.8 &            90.0 &            90.0 \\
                    tae &           82.1 &            84.8 &  \bfseries 85.4 &            82.8 &            79.4 &            82.8 &            82.8 \\
            tic-tac-toe &           69.8 &            81.5 &            84.2 &  \bfseries 92.8 &            88.3 &            90.4 &            90.4 \\
                titanic &           73.4 &            81.0 &            81.6 &  \bfseries 82.0 &            81.5 &            78.8 &            78.8 \\
                   wdbc &           71.5 &  \bfseries 91.4 &            89.1 &            91.2 &            90.4 &            91.0 &            91.0 \\
                   wine &           80.8 &  \bfseries 86.6 &            81.4 &            86.0 &            83.7 &            85.9 &            85.9 \\
\bottomrule
\end{tabular}
\end{table}

\begin{table}
\label{fig:depth=7_metric=training_accuracy}
\caption{Average in-sample accuracies (\%) for each dataset for depth 7}
\begin{tabular}{lrllllll}
\toprule
                   data &  BinOCT &         FlowOCT &      BendersOCT &            CART &           RST-M &           RST-G &          Hybrid \\
\midrule
                  adult &    76.1 &            76.6 &            76.3 &            83.4 &  \bfseries 83.9 &            83.4 &            83.4 \\
       agaricus-lepiota &    72.7 &            92.4 &            96.6 & \bfseries 100.0 & \bfseries 100.0 & \bfseries 100.0 & \bfseries 100.0 \\
          balance-scale &    65.9 &            80.6 &            82.6 &            86.4 &            88.3 &  \bfseries 88.8 &  \bfseries 88.8 \\
banknote-authentication &    72.1 &            84.8 &            94.3 &            92.9 &            96.8 &  \bfseries 96.9 &  \bfseries 96.9 \\
         car-evaluation &    71.6 &            70.0 &            84.1 &            92.6 &            94.4 &  \bfseries 95.2 &  \bfseries 95.2 \\
               diabetes &    70.4 &            71.2 &            72.5 &            80.3 &  \bfseries 86.9 &            84.3 &            84.3 \\
               haberman &    79.1 &  \bfseries 85.8 &            85.7 &            83.8 &            85.6 &            85.4 &            85.4 \\
               kr-vs-kp &    76.1 &            69.0 &            75.6 &            96.7 &            97.7 &  \bfseries 98.4 &  \bfseries 98.4 \\
                monks-1 &    89.6 & \bfseries 100.0 & \bfseries 100.0 &            91.3 & \bfseries 100.0 & \bfseries 100.0 & \bfseries 100.0 \\
                monks-2 &    67.7 &            94.4 &            76.0 &            99.3 &            97.9 & \bfseries 100.0 & \bfseries 100.0 \\
                monks-3 &    88.9 &  \bfseries 99.0 &            98.9 &            98.9 &            96.7 &  \bfseries 99.0 &  \bfseries 99.0 \\
                nursery &    38.6 &            83.4 &            73.0 &            92.2 &            92.9 &  \bfseries 93.6 &  \bfseries 93.6 \\
          seismic-bumps &    93.5 &            93.9 &            93.5 &            95.2 &            95.5 &  \bfseries 96.1 &  \bfseries 96.1 \\
               spambase &    60.9 &            73.8 &            72.8 &            92.4 &  \bfseries 93.7 &            93.3 &            93.3 \\
                    tae &    96.3 &            97.5 &  \bfseries 97.6 &            96.2 &            95.2 &            97.1 &            97.1 \\
            tic-tac-toe &    69.7 &            85.1 &            84.3 &            97.8 &            98.3 &  \bfseries 98.9 &  \bfseries 98.9 \\
                titanic &    76.7 &            85.3 &            82.3 &            87.6 &            88.6 &  \bfseries 89.4 &  \bfseries 89.4 \\
                   wdbc &    75.9 &            59.1 &            93.5 &            97.6 &            98.6 &  \bfseries 98.9 &  \bfseries 98.9 \\
                   wine &    87.2 &  \bfseries 97.1 &  \bfseries 97.1 &            94.8 &            95.2 &            94.8 &            94.8 \\
\bottomrule
\end{tabular}
\end{table}

\begin{table}
\label{fig:depth=7_metric=test_accuracy}
\caption{Average out-of-sample accuracies (\%) for each dataset for depth 7}
\begin{tabular}{llllllll}
\toprule
                   data &         BinOCT &         FlowOCT &     BendersOCT &            CART &           RST-M &           RST-G &          Hybrid \\
\midrule
                  adult &           76.1 &            76.7 &           76.3 &            82.8 &  \bfseries 82.9 &            82.8 &            82.8 \\
       agaricus-lepiota &           69.4 &            91.3 &           96.6 & \bfseries 100.0 & \bfseries 100.0 & \bfseries 100.0 & \bfseries 100.0 \\
          balance-scale &           60.8 &            73.8 &           74.7 &  \bfseries 75.8 &            72.3 &            72.5 &            72.5 \\
banknote-authentication &           72.6 &            83.3 &           93.2 &            92.0 &  \bfseries 95.5 &            94.8 &            94.8 \\
         car-evaluation &           71.3 &            69.9 &           82.4 &            92.1 &            92.0 &  \bfseries 93.9 &  \bfseries 93.9 \\
               diabetes &           69.6 &            65.1 &           68.8 &            69.8 &  \bfseries 71.2 &            70.5 &            70.5 \\
               haberman & \bfseries 69.2 &            65.4 &           68.6 &            66.3 &            66.7 &            65.4 &            65.4 \\
               kr-vs-kp &           75.8 &            67.8 &           74.2 &            96.0 &            97.2 &  \bfseries 97.9 &  \bfseries 97.9 \\
                monks-1 &           89.2 & \bfseries 100.0 &           99.6 &            85.6 & \bfseries 100.0 & \bfseries 100.0 & \bfseries 100.0 \\
                monks-2 &           66.7 &            89.8 &           65.4 &            99.5 &            93.8 & \bfseries 100.0 & \bfseries 100.0 \\
                monks-3 &           88.4 &            98.6 & \bfseries 98.7 &  \bfseries 98.7 &            93.2 &            97.1 &            97.1 \\
                nursery &           41.0 &            84.9 &           73.4 &            92.0 &            92.6 &  \bfseries 93.1 &  \bfseries 93.1 \\
          seismic-bumps & \bfseries 93.3 &            93.1 &           93.2 &            92.0 &            91.9 &            92.1 &            92.1 \\
               spambase &           62.2 &            73.8 &           72.4 &            90.0 &  \bfseries 90.6 &            90.3 &            90.3 \\
                    tae &           80.7 &            82.8 &           82.8 &  \bfseries 84.1 &            80.7 &  \bfseries 84.1 &  \bfseries 84.1 \\
            tic-tac-toe &           67.5 &            83.1 &           81.7 &  \bfseries 94.6 &            89.2 &            90.3 &            90.3 \\
                titanic &           73.8 &  \bfseries 81.3 &           79.0 &            80.5 &            80.2 &            80.4 &            80.4 \\
                   wdbc &           72.9 &            59.1 &           89.4 &  \bfseries 91.0 &            90.0 &            90.9 &            90.9 \\
                   wine &           79.2 &            83.8 &           76.0 &            88.8 &            87.6 &  \bfseries 89.3 &  \bfseries 89.3 \\
\bottomrule
\end{tabular}
\end{table}

\begin{table}
\label{fig:depth=8_metric=training_accuracy}
\caption{Average in-sample accuracies (\%) for each dataset for depth 8}
\begin{tabular}{lrllllll}
\toprule
                   data &  BinOCT &         FlowOCT &      BendersOCT &            CART &           RST-M &           RST-G &          Hybrid \\
\midrule
                  adult &    76.1 &            76.6 &            76.3 &            83.7 &  \bfseries 84.3 &            83.9 &            83.9 \\
       agaricus-lepiota &    72.7 &            92.4 &            98.4 & \bfseries 100.0 & \bfseries 100.0 & \bfseries 100.0 & \bfseries 100.0 \\
          balance-scale &    65.6 &            81.5 &            80.4 &            89.1 &            91.3 &  \bfseries 91.8 &  \bfseries 91.8 \\
banknote-authentication &    72.1 &            91.9 &            93.1 &            96.0 &            98.0 &  \bfseries 98.1 &  \bfseries 98.1 \\
         car-evaluation &    71.6 &            70.0 &            81.5 &            95.0 &            96.4 &  \bfseries 97.0 &  \bfseries 97.0 \\
               diabetes &    69.6 &            57.7 &            72.2 &            82.1 &  \bfseries 89.5 &            86.2 &            86.2 \\
               haberman &    77.5 &  \bfseries 87.3 &            85.6 &            84.9 &            87.0 &            86.8 &            86.8 \\
               kr-vs-kp &    76.1 &            57.1 &            74.5 &            98.6 &            98.2 &  \bfseries 99.1 &  \bfseries 99.1 \\
                monks-1 &    88.9 & \bfseries 100.0 & \bfseries 100.0 &            95.7 & \bfseries 100.0 & \bfseries 100.0 & \bfseries 100.0 \\
                monks-2 &    69.3 &            91.8 &            73.3 &            99.7 &            99.7 & \bfseries 100.0 & \bfseries 100.0 \\
                monks-3 &    86.3 &            98.9 &            98.9 &            98.9 &            97.6 &  \bfseries 99.0 &  \bfseries 99.0 \\
                nursery &    38.6 &            83.4 &            71.9 &            94.4 &            95.1 &  \bfseries 95.9 &  \bfseries 95.9 \\
          seismic-bumps &    93.5 &            93.9 &            93.5 &            95.9 &            96.1 &  \bfseries 96.7 &  \bfseries 96.7 \\
               spambase &    60.9 &            73.8 &            72.3 &            93.4 &            94.5 &  \bfseries 94.7 &  \bfseries 94.7 \\
                    tae &    94.5 &  \bfseries 97.6 &  \bfseries 97.6 &            97.3 &            96.7 &  \bfseries 97.6 &  \bfseries 97.6 \\
            tic-tac-toe &    69.2 &            82.9 &            77.4 &            98.9 &            99.3 &  \bfseries 99.8 &  \bfseries 99.8 \\
                titanic &    76.7 &            86.2 &            81.2 &            88.7 &            89.8 &  \bfseries 90.7 &  \bfseries 90.7 \\
                   wdbc &    75.9 &            52.0 &            93.1 &            98.4 &            99.2 &  \bfseries 99.6 &  \bfseries 99.6 \\
                   wine &    82.2 &            96.9 &  \bfseries 99.1 &            97.9 &            97.4 &            97.1 &            97.1 \\
\bottomrule
\end{tabular}
\end{table}

\begin{table}
\label{fig:depth=8_metric=test_accuracy}
\caption{Average out-of-sample accuracies (\%) for each dataset for depth 8}
\begin{tabular}{llllllll}
\toprule
                   data &         BinOCT &         FlowOCT &      BendersOCT &            CART &           RST-M &           RST-G &          Hybrid \\
\midrule
                  adult &           76.1 &            76.7 &            76.3 &            82.9 &  \bfseries 83.0 &            82.9 &            82.9 \\
       agaricus-lepiota &           69.4 &            91.3 &            98.3 & \bfseries 100.0 & \bfseries 100.0 & \bfseries 100.0 & \bfseries 100.0 \\
          balance-scale &           60.5 &            71.0 &            72.0 &  \bfseries 75.5 &            72.6 &            73.8 &            73.8 \\
banknote-authentication &           72.6 &            90.8 &            91.6 &            94.7 &            96.1 &  \bfseries 96.6 &  \bfseries 96.6 \\
         car-evaluation &           71.3 &            69.9 &            80.6 &            93.7 &            93.9 &  \bfseries 95.0 &  \bfseries 95.0 \\
               diabetes &           69.0 &            57.0 &            67.7 &            69.8 &  \bfseries 70.9 &            70.2 &            70.2 \\
               haberman &           68.0 &  \bfseries 71.0 &            67.6 &            66.7 &            69.2 &            66.1 &            66.1 \\
               kr-vs-kp &           75.8 &            57.5 &            75.4 &            98.4 &            97.5 &  \bfseries 98.5 &  \bfseries 98.5 \\
                monks-1 &           87.1 & \bfseries 100.0 & \bfseries 100.0 &            91.4 & \bfseries 100.0 & \bfseries 100.0 & \bfseries 100.0 \\
                monks-2 &           67.4 &            87.7 &            61.6 &            99.7 &            97.7 & \bfseries 100.0 & \bfseries 100.0 \\
                monks-3 &           85.9 &            98.2 &  \bfseries 98.6 &            98.0 &            93.3 &            97.5 &            97.5 \\
                nursery &           41.0 &            84.9 &            72.4 &            94.0 &            94.8 &  \bfseries 95.3 &  \bfseries 95.3 \\
          seismic-bumps & \bfseries 93.3 &            93.1 &            93.2 &            91.3 &            91.9 &            91.8 &            91.8 \\
               spambase &           62.2 &            73.8 &            72.0 &            90.1 &  \bfseries 90.7 &  \bfseries 90.7 &  \bfseries 90.7 \\
                    tae &           83.4 &            83.4 &            84.1 &            82.8 &            81.4 &  \bfseries 84.8 &  \bfseries 84.8 \\
            tic-tac-toe &           67.2 &            80.6 &            73.9 &  \bfseries 93.5 &            90.6 &            90.6 &            90.6 \\
                titanic &           73.8 &            80.4 &            77.1 &            79.6 &            80.4 &  \bfseries 80.9 &  \bfseries 80.9 \\
                   wdbc &           72.9 &            52.0 &            88.1 &  \bfseries 91.9 &            89.7 &            90.5 &            90.5 \\
                   wine &           72.4 &            87.1 &            79.3 &  \bfseries 89.9 &            89.3 &            89.8 &            89.8 \\
\bottomrule
\end{tabular}
\end{table}
\end{document}